\documentclass{article}
\newcommand{\RETURN}{\STATE \textbf{return}}

\usepackage[square,numbers,compress]{natbib}
\usepackage{xcolor}
\definecolor{lightgreen}{RGB}{20,210,20}
\definecolor{systembrown}{RGB}{139,90,0}
\definecolor{objectblue}{RGB}{0,120,170}
\definecolor{relationred}{RGB}{255,50,0}
\usepackage[citecolor=lightgreen]{hyperref}

\usepackage{subcaption}

\newcommand{\add}[1]{{\color{black}#1}}

\usepackage{amssymb}
\usepackage{array}
\newcolumntype{x}[1]{>{\centering\arraybackslash}p{#1}}
\newcolumntype{y}[1]{>{\raggedright\arraybackslash}p{#1}}
\newcolumntype{z}[1]{>{\raggedleft\arraybackslash}p{#1}}

\usepackage{makecell}
\definecolor{shapecolor}{rgb}{0.0,0.5,0.0}
\usepackage{pifont}
\usepackage[table]{xcolor}
\usepackage{bm}
\usepackage{CJKutf8}
\usepackage{comment}
\usepackage{makecell}
\usepackage{utfsym}
\usepackage{wrapfig}
\usepackage{bm}

\usepackage{tcolorbox}
\usepackage{multirow}
\usepackage{enumitem}
\usepackage{pifont}

\definecolor{navyblue}{HTML}{0071BC}
\definecolor{hotpink}{HTML}{FF0080}
\definecolor{oai-white}{HTML}{FFFFFF}
\definecolor{oai-black}{HTML}{000000}
\definecolor{oai-red}{HTML}{FF4500}
\definecolor{oai-green}{HTML}{51DA4C}
\definecolor{oai-blue}{HTML}{0000FF}
\definecolor{oai-yellow}{HTML}{FFF639}
\definecolor{oai-magenta}{HTML}{FF45FF}
\definecolor{oai-cyan}{HTML}{00FFFF}
\definecolor{oai-orange}{HTML}{FE7600}
\definecolor{oai-violet}{HTML}{8A2BE2}
\definecolor{oai-brown}{HTML}{A0522D}
\definecolor{oai-green-050}{HTML}{F4FFF4}
\definecolor{oai-green-100}{HTML}{E9FFE8}
\definecolor{oai-green-200}{HTML}{D9FFD8}
\definecolor{oai-green-300}{HTML}{C9FFC7}
\definecolor{oai-green-400}{HTML}{A6FFA3}
\definecolor{oai-green-500}{HTML}{7CF178}
\definecolor{oai-green-600}{HTML}{51DA4C}
\definecolor{oai-green-700}{HTML}{3FA93B}
\definecolor{oai-green-800}{HTML}{2D712A}
\definecolor{oai-green-900}{HTML}{193718}
\definecolor{oai-gray-000}{HTML}{FFFFFF}
\definecolor{oai-gray-100}{HTML}{FAFAFA}
\definecolor{oai-gray-200}{HTML}{F5F5F5}
\definecolor{oai-gray-300}{HTML}{E5E5E5}
\definecolor{oai-gray-400}{HTML}{FFB7A4}
\definecolor{oai-gray-500}{HTML}{CDCDCD}
\definecolor{oai-gray-600}{HTML}{A8A8A8}
\definecolor{oai-gray-700}{HTML}{747474}
\definecolor{oai-gray-800}{HTML}{393939}
\definecolor{oai-gray-900}{HTML}{000000}

\definecolor{mygreen}{HTML}{3cb44b}

\definecolor{myred}{HTML}{E33222}
\definecolor{gr}{RGB}{0, 146, 0}

\usepackage[preprint]{neurips_2025}

\usepackage[utf8]{inputenc} 
\usepackage[T1]{fontenc}    

\usepackage{url}            
\usepackage{booktabs}       
\usepackage{amsfonts}       
\usepackage{nicefrac}       
\usepackage{microtype}      
\usepackage[table]{xcolor}
\usepackage{graphicx}
\usepackage{enumitem}
\usepackage{multicol}
\usepackage{multirow}
\usepackage{amsmath}
\usepackage{algorithm} %
\usepackage[noend]{algorithmic} %

\title{\add{Text-Scene: A Scene-to-Language Parsing Framework for 3D Scene Understanding}}

\author{%
  Haoyuan Li, Rui Liu, Hehe Fan, Yi Yang\thanks{Correspongding Author: Yi Yang (e-mail: yangyics@zju.edu.cn)}\\
  College of Computer Science and Technology, Zhejiang University\\
}

\begin{document}

\maketitle

\begin{abstract}
Enabling agents to understand and interact with complex 3D scenes is a fundamental challenge for embodied artificial intelligence systems. While Multimodal Large Language Models (MLLMs) have achieved significant progress in 2D image understanding, extending such capabilities to 3D scenes remains difficult: 1) 3D environment involves richer concepts such as spatial relationships, affordances, physics, layout, and so on, 2) the absence of large-scale 3D vision-language datasets has posed a significant obstacle. In this paper, we introduce \textit{Text-Scene}, a framework that \textit{automatically parses 3D scenes into textual descriptions} for scene understanding. Given a 3D scene, our model identifies object attributes and spatial relationships, and then generates a coherent summary of the whole scene, bridging the gap between 3D observation and language without requiring human-in-the-loop intervention. By leveraging both geometric analysis and MLLMs, Text-Scene produces descriptions that are accurate, detailed, and human-interpretable, capturing object-level details and global-level context. Experimental results on benchmarks demonstrate that our textual parses can faithfully represent 3D scenes and benefit downstream tasks. To evaluate the reasoning capability of MLLMs, we present InPlan3D, a comprehensive benchmark for 3D task planning, consisting of $3174$ long-term planning tasks across $636$ indoor scenes. We emphasize clarity and accessibility in our approach, aiming to make 3D scene content understandable through language. Code and datasets will be released. 
\end{abstract}

\section{Introduction}
\label{introduction}


3D scene understanding is a crucial task of \add{embodied AI} with broad applications in robotics, augmented reality, and autonomous systems~\cite{chen2024end}, which requires agents to perform complex operations in real-world environment~\cite{shridhar2020alfred}.  \add{\textcolor{black}{Previous methods achieved promising accuracy in single 3D tasks, \textit{e.g.} visual grounding and semantic segmentation tasks. However, they lack robust general-understanding capabilities, and their outputs often fail to align with the given language instructions.} Recent studies~\cite{hong20233d, chen2024ll3da, qi2025gpt4scene, huang2023embodied, zheng2024video, huang2023chat, zhu2024unifying} focus on fine-tuning Multimodal Large Language Models (MLLMs) to advance 3D scene understanding, which develop general-purpose assistants. These approaches incorporate the features of detected objects, constructing scene-level 3D representations by integrating multiple techniques: harnessing point-cloud feature or lifting multi-view image features into 3D space. 

The integration of MLLMs with 3D scenes understanding enables MLLMs to describe, reason, and interact in real-word environments. However, bridging 3D scenes and language presents unique challenges: 1) MLLMs are predominantly trained on paired image-text data from the internet, yet the 2D visual knowledge falls short of capturing the complexity inherent in 3D scenes~\cite{zheng2024video}, 2) richly annotated 3D data (\textit{e.g.}, depth maps and point clouds) suitable for fine-tuning MLLMs remain severely scarce~\cite{zheng2025learning},  3) the representation of point clouds or video inputs will bring huge token overhead, which consumes lots of resources and slowing down inference. These limitations constrain the performance potential of MLLMs on tasks, \textit{e.g.} 3D scene understanding and embodied task planning~\cite{jia2024sceneverse}.Given the aforementioned challenges, a fundamental question arises: \add{\textit{Can we propose a more efficient and general solution to interpret 3D scene understanding?}} 
}

In this paper, we introduce a scene-to-language parsing framework, \textbf{Text-Scene},  transforming 3D scenes into structured textual descriptions while efficiently capturing information on geometry. Text-Scene can align 3D geometric features within high-dimensional word-embedding space of MLLMs.  We first parse 3D scenes into two core components: identifying the categories and geometric properties of objects and modeling spatial relationships by forming an intermediate 3D scene graph ($\textbf{\S}$\ref{C3P1} \& $\textbf{\S}$\ref{C3P2}). Then we feed this well-structured 3D information into MLLMs by downstream fine-tuning for MLLMs ($\textbf{\S}$\ref{C3P3}). Through this scene-to-language transformation, MLLMs can generalize across multiple downstream tasks using only a limited amount of 3D data for fine-tuning, eliminating the need for explicit (\textit{e.g.}, point cloud~\cite{huang2023embodied}) or implicit (\textit{e.g.}, neural radiation field~\cite{hong20233d}) spatial representations. Text-Scene offers several advantages: 1) it is compact, reducing memory requirements compared with traditional methods, 2) the textual description is complete and interpretable. Text-Scene achieves strong performance across different downstream tasks with less training costs ($\textbf{\S}$\ref{C3P4}). Our approach achieves state-of-the-art performance on multiple 3D scene understanding datasets, \textit{e.g.} SQA3D~\cite{ma2022sqa3d}, Multi3DRefer~\cite{zhang2023multi3drefer} and ScanRefer~\cite{chen2020scanrefer}. In addition, to comprehensively evaluate embodied task-planning capabilities in 3D scenes, we curate a new benchmark, InPlan3D, consisting of $3,174$ long-term planning tasks across $636$ indoor scenes. Our approach achieves state-of-the-art performance on InPlan3D, compared with the latest 3D MLLMs ($\textbf{\S}$\ref{C4P2}). We provide ablation experiments on modules \textit{e.g.} Sentence Selection and Spatial Reasoning to verify effectiveness ($\textbf{\S}$\ref{C4P3}).

The contributions of this paper are summarized as follows:
\begin{itemize}[leftmargin=*, topsep=0pt, itemsep=2pt, parsep=0pt]
\item  We introduce Text-Scene, an automatic scene-to-language parsing framework. By retaining fine-grained object attributes and spatial relationships, Text-Scene enables seamless adaptation to existing MLLMs for migrating powerful parsing capabilities in the 2D filed to 3D scenes.
\item  To explore the effectiveness of the performance of MLLMs on embodied task planning, we conduct InPlan3D benchmark, consists of $3,174$ tasks across $636$ scenes.
\item Our method achieves state-of-the-art performance across various 3D scene-languange benchmarks. Furthermore, our method could be easily transfered on real-world embodied tasks (\textit{e.g.}, embodied task planning). Extensive ablation experiments on token consumption and training resources highlight the significant advantages in terms of efficiency. 
\end{itemize}

\section{Related Work}

\textbf{3D Scene Understanding.} In the rapidly progressing domain of 3D scene understanding, language has emerged as a powerful tool for conveying contextual cues and formulating user intent. Core tasks including (1) 3D Visual Grounding~\cite{chen2020scanrefer, zhang2023multi3drefer, chen2023end, wang20233drp, zhao20213dvg, wang2023distilling, unal2024four}, which aims to localize target objects in 3D space based on textual descriptions; (2) 3D Question Answering (3D QA)~\cite{azuma2022scanqa, parelli2023clip, ma2022sqa3d}, which addresses scene-level reasoning and information retrieval through natural language queries; (3)  3D Dense Captioning~\cite{chen2021scan2cap, yuan2022x, jiao2022more, chen2023end, chen2024vote2cap, cai20223djcg, chen2022d}, which requires generating fine-grained object-level captions with accurate spatial localization. Traditional models~\cite{zhu20233d, jin2023context} often depend on dedicated task-specific heads, which constrain their flexibility in adapting to open-ended user-assistant interactions and limits their broader applicability in general-purpose multi-modal reasoning.

\textbf{LLMs with 3D Input.} Recent efforts have increasingly focused on integrating 3D scene information into large language models (LLMs) to advance 3D scene understanding~\cite{chen2023end, chen2024vote2cap, chen2024ll3da, fu2024scene, guo2023point, hong20233d, wu2023eda, fan2024navigation, fan2025scene}. 3D-LLM~\cite{hong20233d} initially leverages rendered 2D views as input to LLMs. Methods like Chat3D~\cite{huang2023chat}, LEO~\cite{huang2023embodied}, and ChatScene~\cite{huang2023chatscene} rely on off-the-shelf 3D detectors to generate object proposals, which are then integrated into language models. Meanwhile, GPT4Scene~\cite{qi2025gpt4scene} captures object-level and scene-level semantic features by leveraging multi-view images and rendered BEV images, respectively. Similarly, Video 3D LLM~\cite{zheng2024video} introduces a novel paradigm that implicitly embeds 3D spatial information into video representations, eliminating the need for specialized 3D encoders. However, directly feeding scene point clouds or using multi-view images introduces longer token sequences, resulting in high training costs. Moreover, inconsistencies in cross-modal representations limit the spatial reasoning capability. To address these, we propose an automatic scene-to-language translation framework that accurately captures both object attributes and 3D spatial relationships.

\textbf{Multimodal Embodied Tasks.} The emergence of general-purpose models exhibit strong performance across a wide range of multi-modal tasks~\cite{lu2022unified, lu2024unified, wang2023images, kirillov2023segment, achiam2023gpt, hurst2024gpt, kim2024openvla,liu2023bird,liu2024vision}, while they still face challenges when deployed in embodied scenarios that require perception, planning, and interaction with the environment~\cite{liu2024vol}. To address this gap, various approaches have been proposed~\cite{ahn2022can, huang2022inner, huang2023embodied}. We propose InPlan3D, a more diverse and general-purpose benchmark dataset designed to evaluate the quality of task generation. Tasks in InPlan3D are constructed from the perspective of a indoor service robot, aiming to explore the potential of empowering embodied robot.

\begin{figure}[!t]
    \centering
    \includegraphics[width=\linewidth]{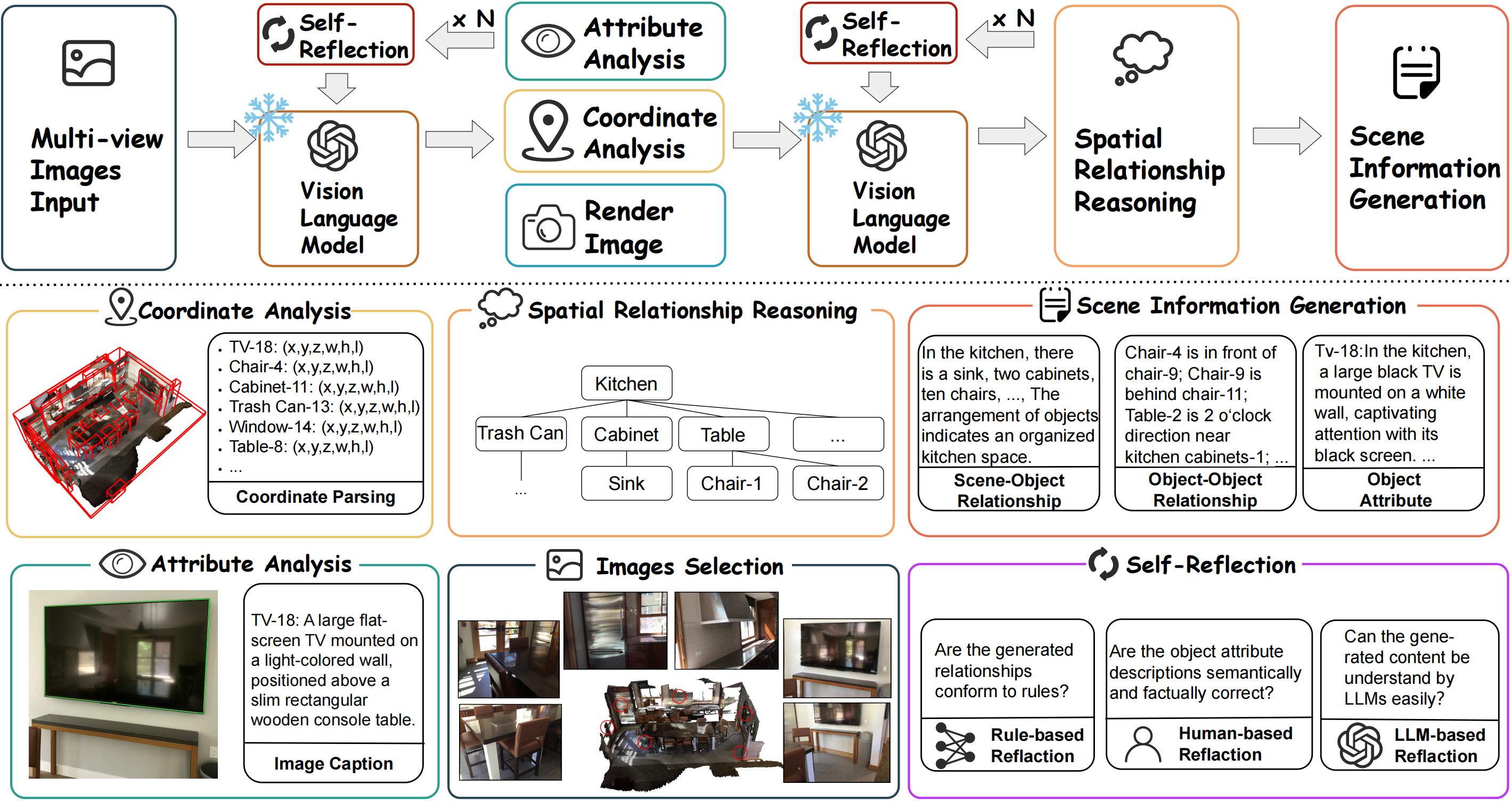}
    \vspace{-5pt}
    \caption{Illustration of Text-Scene, an automatic scene-to-language parsing framework. We first identify objects' categories and geometric properties. Then analysis the salient spatial relationships among them by forming an intermediate 3D scene graph. We insert a self-reflection mechanism between core modules to ensure the generated content is aligned with human intent and adheres to physical laws, providing feedback to the LLMs using different types of evaluation method.}
    \label{fig_scene_graph_generation}
    \vspace{-10pt}
    
\end{figure}

\section{Method}

In this section, we detail the Text-Scene framework. We first introduce the pipeline of the text-driven scene parsing algorithm ($\textbf{\S}$\ref{C3P1}). We further illustrate the scene-to-language generation framework ($\textbf{\S}$\ref{C3P2}). Then, we describe how to align the text-driven scene information with MLLMs ($\textbf{\S}$\ref{C3P3}). Implementation details are provided for reference ($\textbf{\S}$\ref{C3P4}).

\subsection{Scene Parsing}
\label{C3P1}

\textbf{Image Sampling.} Given a raw egocentric video, whose each frame captures a portion of the 3D scene, we first randomly select $n$ frames ${\mathcal{V}}=\{\bm{I}_1, \bm{I}_2, \dots, \bm{I}_n\}$ with corresponding camera extrinsics ${\mathcal{\varepsilon}}=\{\bm{E}_1, \bm{E}_2, \dots, \bm{E}_n\}$. Then we could reconstruct 3D point clouds ${\mathcal{P}}$: ${\mathcal{P}}={\mathcal{R}}(\{(\bm{I}_i,\bm{E}_i)\}_{i=1}^N)$, where ${\mathcal{R}}$ represents images to point cloud projection using 3D reconstruction.

\textbf{Spatial Relationship Reasoning.} To successfully recognize diverse objects within a scene, reason about their spatial relationships, and perform task planning and execution, an agent must possess strong scene semantic understanding capabilities. We could obtain instance masks ${\mathcal{M}}=\{\bm{M}_1,\bm{M}_2,\dots, \bm{M}_K\}$ by applying 3D instance segmentation methods (\textit{e.g.}, Mask3D~\cite{schult2023mask3d}), where $K$ denotes the total number of objects in the scene. As shown in Algorithm~\ref{algorithm_spatial_relationship}, we propose a spatial relationship reasoning framework that integrates geometric proximity, camera-view-based directional inference, and semantic priors to generate fine-grained and interpretable object-level spatial relations. For detailed procedure for relationship computation, please refer Appendix~\ref{appendix_spatial}. Consequently, the spatial relationship among all objects is encoded as a set of node-edge triplets.

\textbf{Instance Projection.} When processing a query like \textit{find a table directly in front of the black armchair}, the model must first identify key attributes such as \textit{black}, \textit{armchair}, and \textit{table}, inferring based on the combined attribute and spatial cues. This capability is fundamental for supporting downstream tasks including object localization, detailed description generation, and high-level task planning. We project the 3D bounding box of a given object onto multi-view images. The corresponding image regions are then cropped based on the projected bounding boxes and processed using BLIP-2~\cite{li2023blip} to generate multiple brief captions for the corresponding object. Next, we apply CLIP~\cite{radford2021learning} to compute the similarity between each cropped image and its brief caption, thereby assessing text–visual alignment. Among all generated captions, we select the top $10$ sentences with the highest CLIP similarity scores. The candidate sentences are subsequently passed to an advanced large language model (\textit{e.g.}, GPT-4o~\cite{hurst2024gpt}) to integrate and refine the outputs from BLIP-2.

\subsection{Scene-to-language Translation}
\label{C3P2}

\textbf{Scene Information Generation.} 
After parsing the attribute and positional relationships of each object, we employ an advanced MLLM (\textit{e.g.}, GPT-4o~\cite{hurst2024gpt}) to generate structured descriptions, which comprises three components: 1) System Message that instructs the LLM about the structure of the inputs; 2) Object Caption section, wherein the attributes of each object are described in language; 3) Relationship Generation component serializes the scene graph, represented as $\{\text{obj}_1, \text{obj}_2, \text{rel}\}$ triplets, into coherent natural language expressions suitable for LLM processing. 

\textbf{Self-Reflection.} As shown in Figure~\ref{fig_scene_graph_generation}, after initially generating object-level captions and inter-object relationships, it is crucial to perform reflective analysis and targeted optimization to ensure the textual content faithfully represents the 3D scene. We observe that direct translations from visual features or coordinate data may occasionally result in incomplete or ambiguous descriptions, especially in spatially dense environments. To address this, we introduce a refinement pipeline that incorporates spatial priors, context-aware consistency checks, and human-in-the-loop feedback where necessary. As detailed in Algorithm~\ref{algorithm_self_reflection}, given the object ID to be evaluated, we first retrieve the corresponding \textbf{Caption} $\mathcal{C}$ and \textbf{Spatial Relationship} $\mathcal{R}$ from the \textbf{Scene Information}. These textual descriptions are then fed into a pretrained \textbf{Value Function} $\mathcal{V}$, which jointly considers the marked multi-view images to assign a quality score. More details are included in Appendix~\ref{S3}. 
\vspace{-0.5em}

\begin{minipage}[t]{0.55\textwidth}
\vspace{-1.5em}
\begin{algorithm}[H]
\small
\caption{Spatial Relationship Reasoning}
\begin{algorithmic}[1]
\REQUIRE Objects $A$, $B$ with centroids
        $\mathbf{p}_A,\mathbf{p}_B\!\in\!\mathbb{R}^{3}$,
        sizes $\mathbf{s}_A,\mathbf{s}_B\!\in\!\mathbb{R}^{3}$;
        camera forward $\mathbf{f}_{\text{cam}}$;
        horizon vector $\mathbf{x}$;
        proximity factor $\beta$, tolerance $\theta_{\text{tol}}$;
        semantic prior $\mathcal{R}_{\text{prior}}$ 
\ENSURE Spatial relationship set $\mathcal{R}_{rel} \in \emptyset$
\IF{$(A,B)\in\mathcal{R}_{\text{prior}}$}
    \STATE \textbf{return} relation from $\mathcal{R}_{\text{prior}}$
\ELSIF{$\Vert \mathbf{p}_A - \mathbf{p}_B \Vert < \beta \times \max(\|\mathbf{s}_A\|,\|\mathbf{s}_B\|)$}
    \STATE Update distance $\mathcal{R}_{rel} \gets \mathcal{R}_{rel} \cup \{\text{nearby}\}$
\ENDIF
    \STATE $\mathbf{r} \leftarrow \mathbf{p}_A - \mathbf{p}_B$ , $v \leftarrow \mathbf{r}\,\cdot\,\mathbf{f}_{\text{cam}}$, $\theta \leftarrow 
        \arccos\!\left(
          \frac{\mathbf{r}\cdot \mathbf{x}}
               {\lVert \mathbf{r}\rVert
                \,\lVert \mathbf{x}\rVert}
        \right)$
    \IF{$v>0$}
        \STATE Update vertical $\mathcal{R}_{rel} \gets \mathcal{R}_{rel} \cup \{\text{is above}\}$
    \ELSE
        \STATE Update vertical $\mathcal{R}_{rel} \gets \mathcal{R}_{rel} \cup \{\text{is below}\}$
    \ENDIF
    \IF{$\theta < \theta_{\text{tol}}$}
        \STATE Update horizontal $\mathcal{R}_{rel} \gets \mathcal{R}_{rel} \cup \{\text{in front of}\}$
    \ELSE
        \STATE Update horizontal $\mathcal{R}_{rel} \gets \mathcal{R}_{rel} \cup \{\text{left of / right of}\}$
    \ENDIF
    \STATE Partition $[0^{\circ},360^{\circ})$ into $N$ sectors, $k \leftarrow \operatorname{sector}(\theta, N)$ 
    \STATE Update angular $\mathcal{R}_{rel} \gets \mathcal{R}_{rel} \cup \ \{k \; \text{o'clock}\}$

\end{algorithmic}
\label{algorithm_spatial_relationship}
\end{algorithm}
\end{minipage}
\hfill
\begin{minipage}[t]{0.44\textwidth}
\vspace{-1.5em}
\begin{algorithm}[H]
\small
\caption{Self-Reflection Mechanism}
\begin{algorithmic}[1]
\REQUIRE Captions $\mathcal{C}$ and relationship $\mathcal{R}$; marked images $\mathcal{I}$;
        advanced LLM $\mathcal{M}$; value function $\mathcal{V}$; threshold $\tau$; GT label $\mathcal{G}$
\ENSURE Refined captions $\hat{\mathcal{C}}$, relationship $\hat{\mathcal{R}}$
\STATE $\hat{\mathcal{C}} \leftarrow \mathcal{C}$;\quad $\hat{\mathcal{R}} \leftarrow \mathcal{R}$
\FORALL{object caption $c_i \in \hat{\mathcal{C}}$}
    \STATE Formulate QA prompt $\mathcal{Q}_i$ from $c_i$
    \STATE Obtain predicted index $o_i \leftarrow \mathcal{M}(\mathcal{Q}_i)$
    \STATE $s_i \leftarrow \operatorname{Accuracy}(o_i, \mathcal{G})$
    \IF{$s_i < \tau$}
        \STATE Obtain correction $c_i^{\prime} \leftarrow \mathcal{V}(c_i)$
        \STATE Update $\hat{\mathcal{C}} \leftarrow (\hat{\mathcal{C}}\setminus\{c_i\}) \cup \{c_i^{\prime}\}$
    \ENDIF
\ENDFOR
\FORALL{relation $r_j \in \hat{\mathcal{R}}$}
    \STATE Compose input pair $(\mathcal{I}, r_j)$
    \STATE $s_j \leftarrow \mathcal{M}\big((\mathcal{I}, r_j)\big)$ 
    \IF{$s_j < \tau$}
        \STATE $r_j^{\prime} \leftarrow \mathcal{V}(r_j)$
        \STATE Update $\hat{\mathcal{R}} \leftarrow (\hat{\mathcal{R}}\setminus\{r_j\}) \cup \{r_j^{\prime}\}$
    \ENDIF
\ENDFOR
\RETURN~$\hat{\mathcal{C}},\; \hat{\mathcal{R}}$
\end{algorithmic}
\label{algorithm_self_reflection}
\end{algorithm}
\end{minipage}

\subsection{3D Scene Understanding with Sentence Selection}
\label{C3P3}

\begin{figure}[!t]
    \centering
    \vspace{-5pt}
    \includegraphics[width=\linewidth]{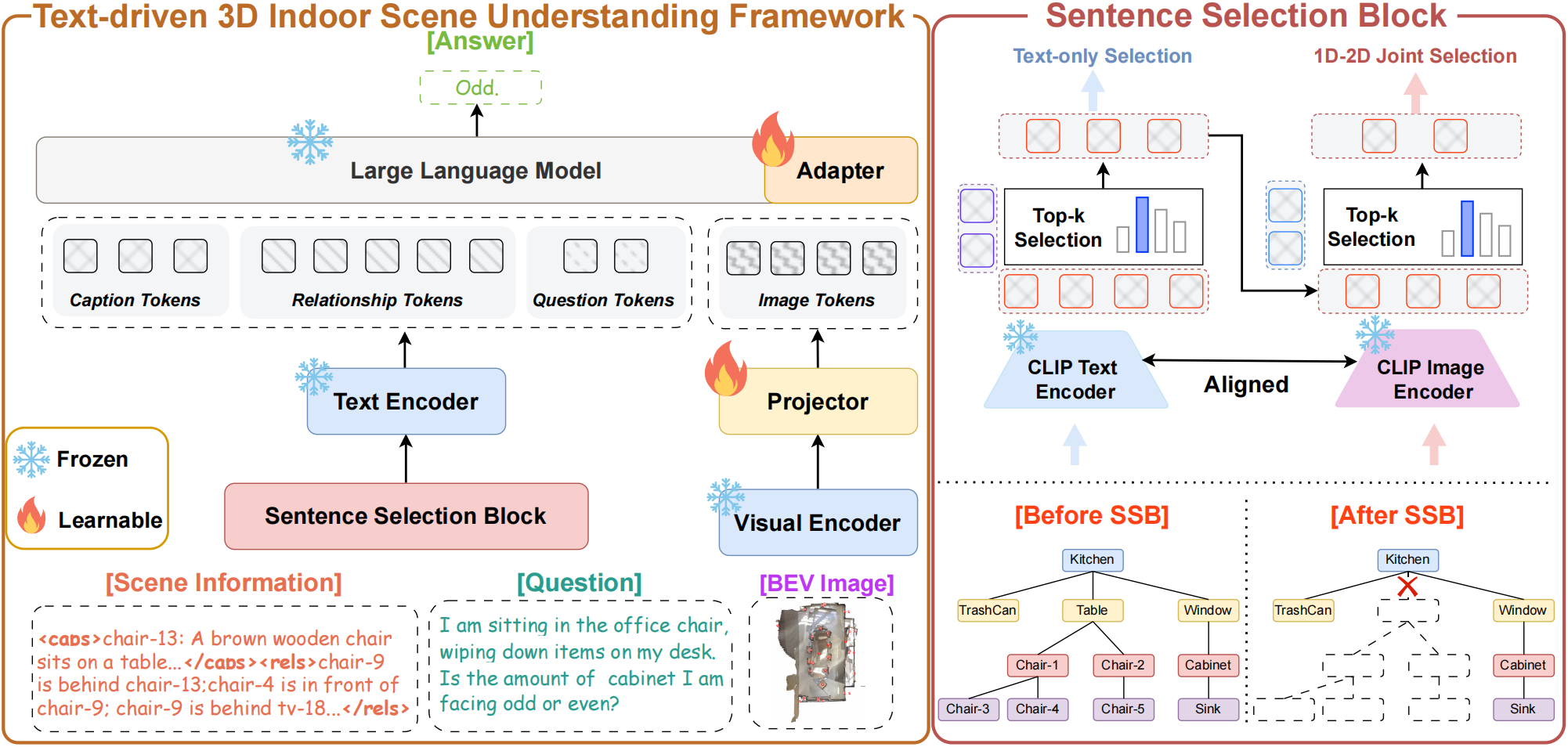}
    \caption{The architecture of the Text-Scene. Given object-level textual captions, the Sentence Selection Block computes token-level similarity scores between the captions and the questions. Then selects the most semantically relevant tokens with respect to the question intent. The combined representation is subsequently fed into the LLM to generate context-aware answers. 
}
    \label{fig_proposed_framework}
    \vspace{-10pt}
\end{figure}

\textbf{Multimodal Reasoning Framework.} Our goal is to extend pre-trained MLLMs for textual scene information inputs. We leverage scene-to-language translation framework for multimodal alignment ($\textbf{\S}$\ref{C3P2}), which parses semantic information and spatial relationship between objects, eliminates the need for heavy multi-modal alignment. As shown in Figure~\ref{fig_proposed_framework}, to enable MLLMs to effectively utilize textual scene information represented, we design the following tasks to facilitate alignment between scene representations and model understanding: 1) \textit{Scene-level caption}: Given the text-driven representation, generate a brief caption about the indoor scene, 2) \textit{Spatial relationship reasoning}: Given the text-driven representation and question, predict the answer. 

\textbf{Sentence Selection Block.} For MLLMs, the textual information in $\textbf{\S}$\ref{C3P2} is overly verbose, which may hinder efficient context comprehension and reasoning. To address this issue, we introduce a sentence selection mechanism that filters out irrelevant content and preserves question-relevant information, thereby enhancing the model's capacity for grounded scene understanding. During the text-only pre-alignment stage, the input consists of a set of object captions from scene information with related questions. These textual components are first encoded using a frozen CLIP Text Encoder within the Sentence Selection Block, which represents each sentence as a sequence of token embedding. Then computes the cosine similarity between the question embedding and each caption embedding to assess their semantic relevance. Based on the computed similarity scores, the top-$k$ most relevant caption tokens are selected. Detailed computation process is shown in Equation~\ref{equa_top_k}:
\vspace{-5pt}
\begin{equation}
\tilde{\bm{Q}} = \frac{\bm{Q}}{\|\bm{Q}\|}, \tilde{\bm{C}} = \frac{\bm{C}}{\|\bm{C}\|}, \bm{S}=\tilde{\bm{C}}\cdot\tilde{\bm{Q}}^\top ,\boldsymbol{\alpha}_{i} = \text{Softmax}(\sum_{j=1}^m\bm{S}_{ij}\cdot\bm{C}_{ij}),
\label{equa_top_k}
\end{equation}
\vspace{-10pt}

where $\bm{Q} \in \mathbb{R}^{n \times d}$ is the question token embeddings, $\bm{C} \in \mathbb{R}^{m \times d}$ is the  caption token embeddings, $n$ and $m$ is the number of question and caption tokens, respectively; $\tilde{\bm{Q}}, \tilde{\bm{C}}$ is L2-normalized token embeddings, $\bm{S} \in \mathbb{R}^{m \times n}$ is pair-wise cosine similarity matrix between caption and question tokens, $\boldsymbol{\alpha} \in \mathbb{R}^m$ is normalized attention weights over caption tokens. The top-$k$ caption tokens with the highest $\bm{\alpha}_i$ scores are selected as the most semantically relevant context for the question. During the fine-tuning phase, we repeat the above process by replacing the matrix $\bm{Q}$ with image feature embeddings extracted by the CLIP Image Encoder, and replacing $\bm{C}$ with the caption token embeddings obtained from the first-round selection. The image features are then used to further refine and filter the caption tokens to better align with the scene context. After passing through the Sentence Selection Block, we can select the top-$k$ objects and their corresponding captions from the original pool of over $60$ objects and captions, reducing token consumption. Also, we can also improve the performance by providing BEV images~\cite{qi2025gpt4scene}. Following the pretraining-finetuning paradigm, the pre-aligned MLLMs can be fine-tuned with BEV images' input for improved downstream task performance. Comparative experiment in Appendix~\ref{S1} shows that BEV-assisted approach can improve performance in QA and Caption tasks, while the improvement in grounding tasks is limited.

\textbf{Loss function.} To standardize the training process, all tasks are reformulated into a unified user-assistant interaction format. Consequently, during the joint training phase, the model is optimized solely using the Cross-Entropy loss from the language modeling objective. The goal is to learn the trainable parameters $\theta$ by minimizing the negative log-likelihood of the assistant’s target response textual sequence $\bm{t}^\text{res}$. Given the input prefix textual sequence $\bm{t}^\text{prefix}$ which includes system messages, the top-k selected scene information and user instructions, the loss function is defined in Equation~\ref{equal_loss_function}:
\vspace{-5pt}
\begin{equation}
    \mathcal{L}(\theta) = - \sum_{i=1}^{k} \log P\left(\bm{t}_i^{\text{res}} \mid \bm{t}_{[1, \dots, i-1]}^{\text{res}}, \bm{t}^{\text{prefix}} \right),
    \label{equal_loss_function}
\end{equation}
\vspace{-10pt}

where $k$ is the number of tokens in the response sequence, and $\bm{t}_{[1,\dots,i-1]}^{\text{res}}$ denotes the sequence of the previous $i-1$ tokens in the response. The set of trainable parameters $\theta$ represents the visual projector, a 3-layer-MLP, as long as LLM adapter.



\subsection{Implementation Details}
\label{C3P4}

We leverage Qwen2-7B as the LLM backbone. For better performance, we utilize the text tokenizer and the ViT~\cite{wang2023images} from Qwen2-VL~\cite{wang2024qwen2} as the multi-modal encoder. In Sentence Selection Block, we use pretrained CLIP-ViT-L~\cite{radford2021learning} as the multi-modal encoder. We use LoRA~\cite{hu2022lora} for supervised fine-tuning with a rank of $8$. During training, the text tokenizer, ViT, CLIP model and LLM backbone are frozen, and the projector and additional adapter for LLM is trainable. We train the model on a mixture of tasks comprising scene-level caption and spatial relationship reasoning with the ratio of 1:3. Text-Scene is trained on four 80G-A800 GPUs in $13$ hours. More training details of Text-Scene and baselines are included in Appendix~\ref{S5}.

\begin{figure}[!t]
    \centering
    \vspace{-5pt}
    \includegraphics[width=\linewidth]{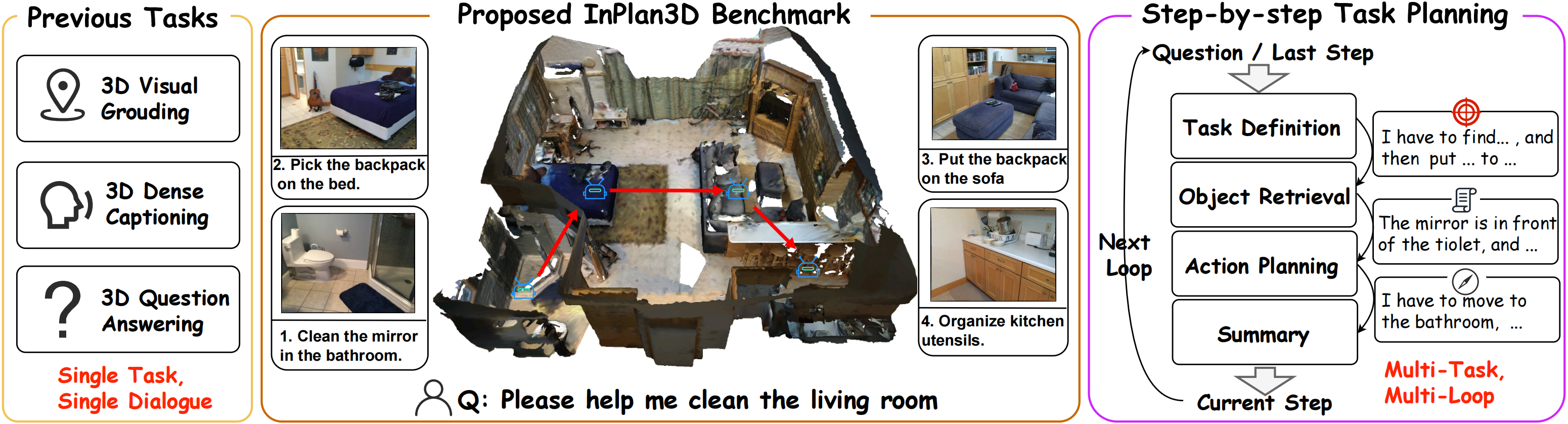}
    \caption{Illustration of proposed InPlan3D benchmark. InPlan3D benchmark emphasizes the model's ability to solve problems step-by-step, requiring problem define, object retrieval, and action planning abilities. Differ from previous tasks, InPlan3D focuses on evaluating reasoning abilities, pushing forward the development of general-purpose, human-aligned embodied agents.}
    \label{fig_inplan3d}
    \vspace{-10pt}
\end{figure}

\section{Experiment}
Text-Scene enables the LLM to perform various 3D indoor scene understanding tasks. Firstly, we present a detailed introduction to the datasets used for evaluation, the metrics employed to assess model performance ($\textbf{\S}$\ref{C4P1}). Then we evaluate Text-Scene on various downstream tasks (\textit{e.g.}, 3D QA, 3D visual grounding, 3D embodied planning) with both text-only setting and normal setting ($\textbf{\S}$\ref{C4P2}). We conduct a series of ablation studies to demonstrate the effectiveness of the Spatial Reasoning algorithm and Sentence Selection block , and we further discuss its advantages over current state-of-the-art methods in terms of token efficiency and overall inference performance ($\textbf{\S}$\ref{C4P3}).

\subsection{Experimental Setting} 
\label{C4P1}
\textbf{Datasets.} We conduct experiments on six different benchmarks across $1,513$ scenes: ScanQA~\cite{azuma2022scanqa} and SQA3D~\cite{ma2022sqa3d} for visual question answering, Scan2Cap~\cite{chen2021scan2cap} for dense captioning, ScanRefer~\cite{chen2020scanrefer} for single-object visual grounding, and Multi3DRefer~\cite{zhang2023multi3drefer} for multi-object visual grouding. In addition, we propose InPlan3D, a benchmark to evaluate the model's capability in indoor task planning based on ScanNet~\cite{dai2017scannet}. In Figure~\ref{fig_inplan3d}, existing benchmarks (\textit{e.g.}, 3D Quesion Answering, 3D Visual Grounding) mainly focus on specific tasks, with single-turn dialogue format. In contrast, InPlan3D incorporates multi-task and multi-turn reasoning dialogue, requiring the model to understand and reason over complex environments (see more details in Appendix~\ref{S4}).

\textbf{Metrics.} Following existing methods~\cite{huang2023embodied,huang2023chat,qi2025gpt4scene}, we assess accuracy using Acc@0.25 and Acc@0.5 for ScanRefer~\cite{chen2020scanrefer} with IoU thresholds of 0.25 and 0.5. For Multi3DRefer~\cite{zhang2023multi3drefer}, we employ a F1 score at IoU thresholds of 0.25 and 0.5. For Scan2Cap~\cite{chen2021scan2cap}, we utilize CIDEr@0.5 and BLEU-4@0.5. For ScanQA~\cite{azuma2022scanqa}, CIDEr~\cite{vedantam2015cider} and BLEU-4~\cite{papineni2002bleu} are used. SQA3D~\cite{ma2022sqa3d} is evaluated using exact match accuracy (EM) and its refined version, EM-R, for better performance evaluation.

\begin{table}[t]
\centering
\setlength\tabcolsep{5pt}
\caption{\textbf{Comparison with baselines.} \textit{Task-specific Models} are customized for specific tasks through task heads. Point and Vision Encoders correspond to input modalities.}

\resizebox{\linewidth}{!}{
{
\begin{tabular}{lccccccccccc}
\toprule 
\multirow{2}{*}{Method} & \multirow{2}{*}{\shortstack{Point\\Encoder}} & \multirow{2}{*}{\shortstack{Vision\\Encoder}} & \multicolumn{2}{c}{ScanRefer} & \multicolumn{2}{c}{Multi3DRef} & \multicolumn{2}{c}{Scan2Cap}  & \multicolumn{2}{c}{ScanQA} & \multicolumn{1}{c}{SQA3D} \\ 
\cmidrule(lr){4-5} \cmidrule(lr){6-7} \cmidrule(lr){8-9} \cmidrule(lr){10-11} \cmidrule(lr){12-12}
& & & \scalebox{0.85}[1]{Acc@0.25} & \scalebox{0.85}[1]{Acc@0.5} & \scalebox{0.9}[1]{F1@0.25} & \scalebox{0.9}[1]{F1@0.5} & \scalebox{0.85}[1]{B-4@0.5} & \scalebox{0.85}[1]{C@0.5} & C & EM & EM \\
\midrule
\multicolumn{12}{l}{\textit{\textbf{Task-specific Models}}} \\
ScanRefer~\cite{chen2020scanrefer} & \ding{51} & \ding{55} & 37.3 & 24.3 & -- & -- & -- & -- & -- & -- & -- \\
MVT~\cite{huang2022multi} & \ding{51} & \ding{55} & 40.8 & 33.3 & -- & -- & -- & -- & -- & -- & -- \\
3DVG-Trans~\cite{zhao20213dvg} & \ding{51} & \ding{55} & 45.9 & 34.5 & -- & -- & -- & -- & -- & -- & -- \\
ViL3DRel~\cite{chen2022language} & \ding{51} & \ding{55} & 47.9 & 37.7 & -- & -- & -- & -- & -- & -- & -- \\
M3DRef-CLIP~\cite{zhang2023multi3drefer} & \ding{51} & \ding{55} & 51.9 & 44.7 & 42.8 & -- & 38.4 & -- & -- & -- & -- \\
Scan2Cap~\cite{chen2021scan2cap} & \ding{51} & \ding{55} & -- & -- & -- & -- & 22.4 & 35.2 & -- & -- & -- \\
ScanQA~\cite{azuma2022scanqa} & \ding{51} & \ding{55} & -- & -- & -- & -- & -- & -- & 64.9 & 21.1 & 47.2 \\
3D-VisTA~\cite{zhu20233d} & \ding{51} & \ding{55} & 50.6 & 45.8 & -- & -- & 34.0 & 66.9 & 69.6 & 22.4 & 48.5 \\
\midrule
\multicolumn{12}{l}{\textit{\textbf{3D LLMs}}} \\
3D-LLM(Flamingo)~\cite{hong20233d} & \ding{51} & \ding{51} & 21.2 & -- & -- & -- & -- & -- & 59.2 & 20.4 & -- \\
3D-LLM(BLIP2-flant5)~\cite{hong20233d} & \ding{51} & \ding{51} & 30.3 & -- & -- & -- & -- & -- & 69.4 & 20.5 & -- \\
Chat-3D~\cite{wang2023chat} & \ding{51} & \ding{55} & -- & -- & -- & -- & -- & -- & 53.2 & -- & -- \\
Chat-3D v2~\cite{huang2023chatv2} & \ding{51} & \ding{55} & 42.5 & 38.4 & 45.1 & 41.6 & 31.8 & 63.9 & 87.6 & -- & 54.7 \\
LL3DA~\cite{chen2024ll3da} & \ding{51} & \ding{55} & -- & -- & -- & -- & 36.0 & 62.9 & 76.8 & -- & -- \\
SceneLLM~\cite{fu2024scene} & \ding{51} & \ding{55} & -- & -- & -- & -- & -- & -- & 80.0 & 27.2 & 53.6 \\
LEO~\cite{huang2023embodied} & \ding{51} & \ding{51} & -- & -- & -- & -- & 38.2 & 72.4 & 101.4 & 24.5 & 50.0 \\
Grounded 3D-LLM~\cite{chen2024grounded} & \ding{51} & \ding{55} & 47.9 & 44.1 & 45.2 & 40.6 & 35.5 & 70.6 & 72.7 & -- & -- \\
PQ3D~\cite{zhu2024unifying} & \ding{51} & \ding{51} & 57.0 & 51.2 & -- & 50.1 & 36.0 & 80.3 & -- & -- & 47.1 \\
ChatScene~\cite{huang2023chatscene} & \ding{51} & \ding{51} & 55.5 & 50.2 & 57.1 & 52.4 & 36.3 & 77.1 & 87.7 & 21.6 & 54.6 \\
LLaVA-3D~\cite{zhu2024llava} & \ding{51} & \ding{51} & 54.1 & 42.4 & -- & -- & 41.1 & 79.2 & 91.7 & 27.0 & 55.6 \\
GPT4Scene~\cite{qi2025gpt4scene}& \ding{55} & \ding{51} & 62.6 & 57.0 & 64.5 & 59.8 & 40.6 & 79.1 & 96.3 & 26.5 & 60.6 \\
Video-3D LLM~\cite{zheng2024video} & \ding{55} & \ding{51} & 58.1 & 51.7 & 58.0 & 52.7 & 41.3 & 83.8 & \textbf{102.1} & \textbf{30.1} & 58.6 \\
\rowcolor{lightgray!40} Text-Scene (text-only)& \ding{55} & \ding{55} & 59.3 & 53.6 & 63.1 & 58.7 & 36.8 & 80.0 & 89.5 & 22.9 & 57.7 \\
\rowcolor{lightgray!60} Text-Scene  & \ding{55} & \ding{51} & \textbf{64.5} & \textbf{59.4} & \textbf{66.1} & \textbf{60.7} & \textbf{41.7}  & \textbf{84.1} & 93.7 & 23.4 & \textbf{61.2} \\
\bottomrule
\end{tabular}
}
}
\label{tab_3d_llm_performance}
\end{table}

\subsection{Comparison with State-of-the-art Methods}
\label{C4P2}
\textbf{Baselines.} Based on the architecture, baselines can be categorized into task-specific models and 3D LLMs. Traditional task-specific models are typically designed for individual tasks and require separate training on corresponding datasets. In contrast, 3D LLMs are generally capable of handling multiple indoor scene understanding tasks simultaneously. 3D LLMs are trained on various datasets covering diverse tasks, eliminating task-specific design or fine-tuning for each individual task.
\begin{itemize}[leftmargin=*, topsep=0pt, itemsep=2pt, parsep=0pt]
\item{\textbf{Task-Specific Models:}} Models such as ScanRefer~\cite{chen2020scanrefer} and ScanQA~\cite{azuma2022scanqa} establish initial benchmarks for the ScanRefer and ScanQA datasets, respectively. 3D-VisTA~\cite{zhu20233d} aim to develop versatile 3D visual-language frameworks by focusing on pre-training strategies for 3D scene-language alignment. M3DRef-CLIP~\cite{zhang2023multi3drefer} introduces multi-object grounding, enhancing single-object grounding performance. ConcreteNet~\cite{unal2024four}, the state-of-the-art model on ScanRefer~\cite{chen2020scanrefer}, innovates three methods to augment verbal-visual fusion for 3D visual grounding. 

\item{\textbf{3D LLMs:}} 3D-LLM~\cite{hong20233d} utilizes location tokens for object grounding but is constrained by data scarcity. LL3DA~\cite{chen2024ll3da} and SceneLLM~\cite{fu2024scene} processes point clouds directly, responding to textual instructions and visual prompts. Grounded 3D-LLM~\cite{chen2024grounded}, PQ3D~\cite{zhu2024unifying}, and LLaVA-3D~\cite{zhu2024llava} achieve strong performance on 3D visual grounding tasks by joint training with a 3D detection module. Chat3D~\cite{huang2023chat}, LEO~\cite{huang2023embodied} and ChatScene~\cite{huang2023chatscene} integrate visual and point cloud modalities, using language as guidance to facilitate cross-modal fusion and understanding.  GPT4Scene~\cite{qi2025gpt4scene} apply multi-view images and marked BEV images as input, while Video-3D LLM~\cite{zheng2024video} uses videos.
\end{itemize}

\textbf{Performance on Scene Understanding.} In Table~\ref{tab_3d_llm_performance}, Text-Scene outperforms all existing task-specific models and 3D LLM baselines on most tasks, demonstrating the strength of our text-driven framework for 3D scene understanding. The term \textit{text-only} indicates that MLLM has only textual description as input. Compared with current state-of-the-art 3D LLMs that rely heavily on visual inputs, such as GPT4Scene~\cite{qi2025gpt4scene} and Video-3D LLM~\cite{zheng2024video}, our method requires significantly fewer image inputs with superior performance across downstream tasks. For visual grounding, Text-Scene achieves a new state-of-the-art with 64.5\% Acc@0.25 and 59.4\% Acc@0.5 on ScanRefer~\cite{chen2020scanrefer}, and 60.7\% F1@0.5 on Multi3dRefer~\cite{zhang2023multi3drefer}, verifying the efficacy of our text-driven scene parsing and object selection strategies. For 3D VQA tasks, our model attains top performance on both 91.3\% CIDEr and 24.6\% EM on ScanQA and 61.2\% EM on SQA3D, confirming the model's ability to understand, ground, and reason over complex 3D scenes without task-specific fine-tuning.

\textbf{Performance on InPlan3D.} In Table~\ref{tab_task_planning}, we report the performance of the 3D LLMs using the same splits and initial conditions. The term \textit{text-only} indicates that MLLM has only textual description as input. More calculation details of $G_\text{Acc}$ and $T_\text{Acc}$ are shown in Appendix~\ref{S4}. Text-Scene achieves state-of-the-art performance across multiple evaluation dimensions. Despite using only a single BEV image as visual input, significantly fewer than prior 3D LLMs that depend on large-scale multi-view inputs, Text-Scene surpasses most baselines in both task-level and step-level accuracy. Specifically, Text-Scene achieves the highest $G_\text{Acc}$ (47.23\%) and $T_\text{Acc}$ (65.91\%), demonstrating strong task grounding and execution ability. Furthermore, in terms of language similarity, our method leads all competitors with top scores across all metrics. These results reflect the model's superior capability to generate coherent, grounded, and high-quality natural language plans. Notably, Text-Scene even outperforms vision-heavy methods like GPT4Scene~\cite{qi2025gpt4scene} and ChatScene~\cite{huang2023chatscene}.

\begin{table}[h]
\centering
\renewcommand{\arraystretch}{1} 
\setlength\tabcolsep{5pt} 
\caption{\textbf{Comparison on Planning Task.} {$G_\text{Acc}$} evaluates task-level accuracy, while {$T_\text{Acc}$} reflects step-level accuracy, and language quality is measured by METEOR~\cite{banerjee2005meteor}, ROUGE~\cite{lin2004rouge}, BLEU-4~\cite{papineni2002bleu}, and CIDEr~\cite{vedantam2015cider}. }
\resizebox{\linewidth}{!}{
{
\begin{tabular}{lccccccll}
\toprule
\multirow{2}{*}{Method} & \multirow{2}{*}{\shortstack{Point\\Encoder}} & \multirow{2}{*}{\shortstack{Vision\\Encoder}} & \multicolumn{1}{c}{Task-Level} & \multicolumn{1}{c}{Step-Level} & \multicolumn{4}{c}{Language Similarity} \\
\cmidrule(lr){4-4} \cmidrule(lr){5-5} \cmidrule(lr){6-9}
& & & $G_\text{Acc}$ & $T_\text{Acc}$ & METEOR & ROUGE & BLEU-4 & CIDEr \\
\midrule
PQ3D~\cite{huang2023embodied} & \ding{51} & \ding{51} & 36.47 & 46.83 & 12.87 & 38.75 & 15.03 & 70.23 \\
LEO~\cite{huang2023embodied} & \ding{51} & \ding{51} & 37.13 & 47.59 & 13.06 & 39.42 & 15.37 & 71.91 \\
ChatScene~\cite{huang2023chatscene} & \ding{51} & \ding{51} & 38.32 & 48.87 & 13.33 & 40.14 & 15.78 & 72.36 \\
GPT4Scene~\cite{qi2025gpt4scene} & \ding{55} & \ding{51} & 41.52 & 52.45 & 13.98 & 42.28 & 16.87 & 76.71 \\
\rowcolor{lightgray!40} Text-Scene (text-only)& \ding{55} & \ding{55} & 45.69 & 58.28 & 13.79 & 41.66 & 19.12 & 74.31 \\
\rowcolor{lightgray!60} Text-Scene & \ding{55} & \ding{51} & \textbf{47.23} & \textbf{65.91} & \textbf{15.04} & \textbf{44.96} & \textbf{19.87} & \textbf{80.17} \\
\bottomrule
\end{tabular}
}
}
\label{tab_task_planning}
\end{table}

\begin{figure}[!t]
    \centering
    \includegraphics[width=\linewidth]{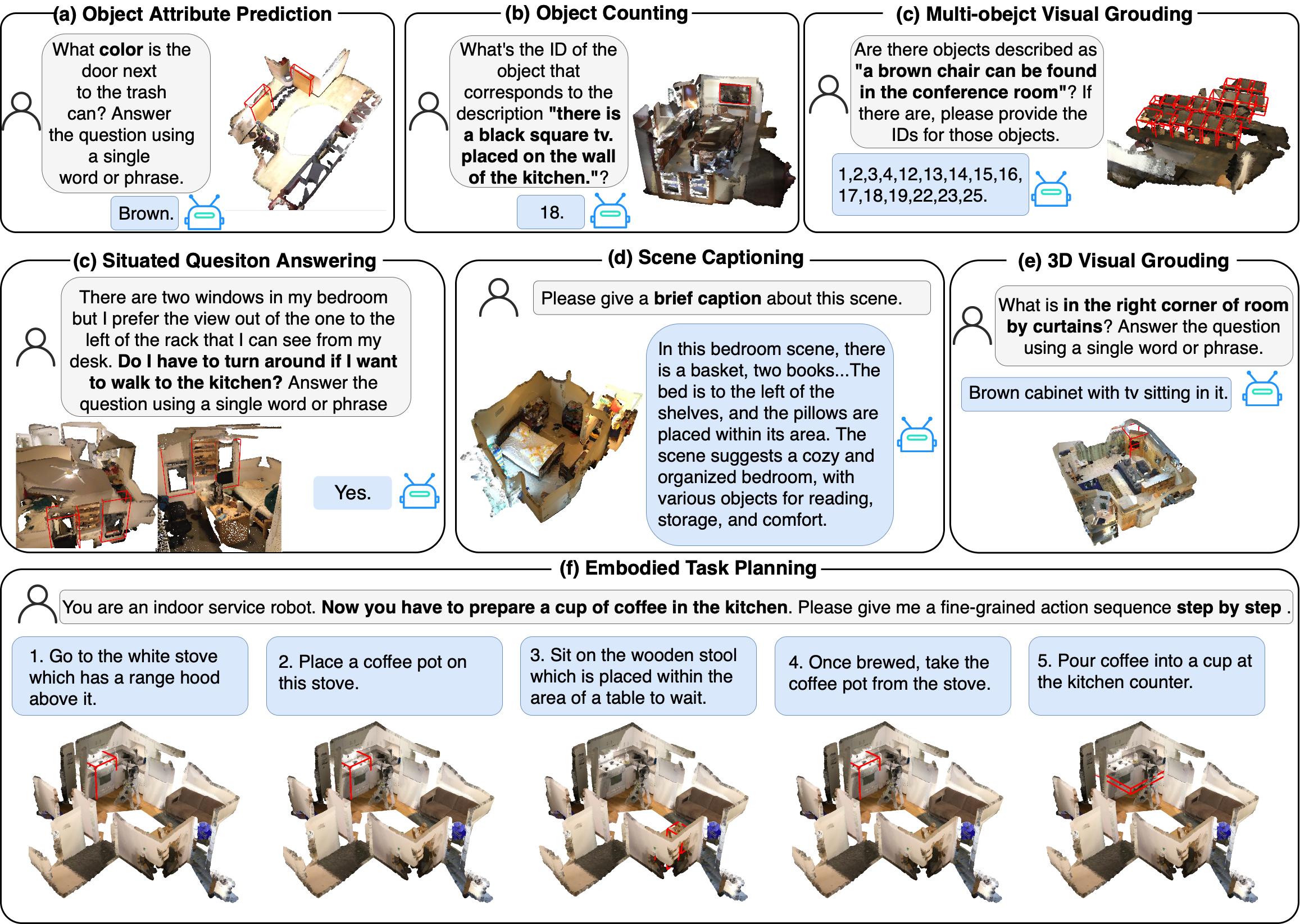}
    \vspace{-10pt}
    \vspace{-5pt}
    \caption{Visualization of various downstream tasks, including 3D dense captioning, 3D question answering (QA), 3D visual grounding, and embodied task planning. Notice that white boxes indicate user instructions, while green boxes present the responses generated by Text-Scene. The rendered scene images are provided for illustrative purposes only and are not directly used as input.}
    \label{fig_multi-tasks}

    \vspace{-10pt}
    
\end{figure}

\newpage

\subsection{Ablation Study}
\label{C4P3}

\vspace{-10pt}
\begin{wraptable}{r}{0.45\textwidth}
\vspace{-5pt}
\caption{\textbf{Ablation study on Relationship Generation.} 
We compare different strategies for generating relational information. 
``Coordinate'' is based on coordinates, ``Simple'' uses semantic relations, and ``Complex'' includes hierarchical and contextual relations.}
\setlength\tabcolsep{6pt}
\renewcommand{\arraystretch}{1.0}  
\resizebox{\linewidth}{!}{
\begin{tabular}{cccc}
\toprule
\multirow{2}{*}{Expression Type} & \multicolumn{1}{c}{ScanRefer} & \multicolumn{1}{c}{Multi3DRef} & \multicolumn{1}{c}{SQA3D} \\
\cmidrule(lr){2-2}\cmidrule(lr){3-3}\cmidrule(lr){4-4}
& Acc@0.5 & F1@0.5 & EM \\
\midrule
Coordinate & 18.4 & 14.4 & 16.5 \\ 
Simple     & 38.9 & 34.3 & 39.6 \\
Complex    & \textbf{59.4} & \textbf{60.7} & \textbf{61.2} \\
\bottomrule
\end{tabular}}
\vspace{-10pt}
\label{tab_ablation_relationship}
\end{wraptable}

\textbf{Effectiveness of scene parsing algorithm.} In Table~\ref{tab_ablation_relationship}, we explore the impact of different relationship generation strategies on model performance. The Coordinate setting directly encodes each object’s 3D center coordinates and bounding box dimensions into textual descriptions, serving as a low-level representation. The Simple Relationship strategy expresses coarse spatial relations such as \textit{in front of}, \textit{behind}, \textit{left of}, or \textit{above}, based solely on the camera view, without precise angular reasoning. In contrast, the Complex Relationship strategy grains angular calculations and describes spatial relations with higher precision, accounting for various directional expressions such as \textit{to the right}, \textit{below} or \textit{at 4 o'clock} thereby capturing multiple plausible spatial configurations. The Complex Relationship setting achieves consistent performance improvements across different benchmarks, with less than $2\times$  increase in input text length.

\textbf{Effectiveness of Sentence Selection Block.} We validate the effectiveness of Sentence Selection block in Table~\ref{tab_ablation_selection}. When no selection is applied and all captions and relationships are fed into the model (All Captions \& Relationships), the performance on downstream tasks is clearly limited. By applying a 1-Round Top-k Selection, which filters scene information based on the input question, the model achieves notable improvements across all benchmarks, while also significantly reducing average inference time to $108$ms, significantly lower than GPT4Scene ($562$ms) and Video-3D LLM ($1204$ms). Further improvement is observed with the 2-Round Top-k Selection, which incorporates BEV image features into a second round of selection. This setup leads to the best performance on all tasks, including ScanRefer Acc@0.5 (59.4), Multi3DRef F1@0.5 (60.7), Scan2Cap(84.1), CIDEr@0.5 (93.7), and SQA3D EM (61.2). These results confirm that combining multi-modal information through a cascaded selection strategy can significantly boost scene understanding and reasoning capabilities, while maintaining a favorable balance between performance and efficiency.
\vspace{-5pt}
\begin{table}[h]
\renewcommand{\arraystretch}{1} 
\setlength\tabcolsep{8pt} 
\vspace{-5pt}
\caption{\textbf{Ablation study on Sentence Selection.} 
\textit{All Captions \& Relationships} denotes that filtering is not applied to the Scene Information.
\textit{1-Round Top-k Selection} and \textit{2-Round Top-k Selection} filters Scene Information once based on textual input or additional BEV images. We also list the performance of GPT4Scene~\cite{qi2025gpt4scene} and Video-3D LLM~\cite{zheng2024video} under default settings for reference.
}
\resizebox{\linewidth}{!}{
\begin{tabular}{ccccccc}
\toprule
\multirow{2}{*}{Method}  & \multirow{2}{*}{\shortstack{Average\\Inference Time}} & \multicolumn{1}{c}{ScanRefer} & \multicolumn{1}{c}{Multi3DRef} & \multicolumn{1}{c}{Scan2Cap} & \multicolumn{1}{c}{ScanQA} & \multicolumn{1}{c}{SQA3D} \\
\cmidrule(lr){3-3} \cmidrule(lr){4-4} \cmidrule(lr){5-5} \cmidrule(lr){6-6} \cmidrule(lr){7-7} 
 & &  Acc@0.5 & F1@0.5 & C@0.5 & CIDEr & EM \\
\midrule
All Captions \& Relationships & 285 ms & 47.2 & 39.6 & 69.1 & 74.9 & 43.7 \\ 
1-Round Top-$k$ Selection     & \textbf{108 ms}& 53.6 & 58.7 & 80.0 & 89.5 & 57.7 \\
2-Round Top-$k$ Selection     & 169 ms& \textbf{59.4} & \textbf{60.7} & \textbf{84.1} & 93.7 & \textbf{61.2} \\
\midrule
GPT4Scene & 962 ms& 57.0 & 59.8 & 79.1 & 96.3 & 60.6 \\
Video-3D LLM & 1204 ms& 51.7 & 52.7 & 83.8 & \textbf{102.1} & 58.6 \\
\bottomrule
\end{tabular}}
\vspace{-15pt}
\label{tab_ablation_selection}
\end{table}

\textbf{Visualization.}
Figure~\ref{fig_multi-tasks} showcases the versatile capabilities of the proposed Text-Scene framework across a wide range of downstream 3D scene-language tasks.  Importantly, the rendered scene images are used for visualization only and are not provided as input to the model, emphasizing the strength of Text-Scene in understanding and reasoning over scenes using text-driven representations alone.

\section{Conclusion}
In this paper, we propose \textit{Text-Scene} that bridges 3D observation and language through automatic scene-to-text parsing. By identifying key objects, attributes, and spatial relationships, our method generates rich, natural-language summaries of 3D scenes without human intervention. This text-driven representation enables holistic 3D scene understanding using only textual input, significantly reducing the reliance on dense visual data and domain-specific encoders. Experiments show that the generated descriptions are accurate, interpretable, and effective for supporting downstream tasks, such as 3D visual grounding, question answering, and dense captioning. Furthermore, to evaluate the practicality in real-world scenarios, we introduce InPlan3D, a diverse benchmark for embodied task planning in indoor environments. The results highlight the potential of leveraging language as a universal medium for 3D scene understanding, offering a scalable and efficient solution.

\newpage

\small
\bibliographystyle{unsrt}
\bibliography{main}








\newpage
\appendix

\section{Additional Ablation Study}
\label{S1}
In this section, we provide additional ablation experiments to further investigate key components of our approach. We compare the reconstruction quality of different point cloud reconstruction methods ($\textbf{\S}$\ref{S1C1}). We also analyze the impact of using pre-annotated labels from the ScanNet dataset \textit{vs.} labels generated by Mask3D on overall performance ($\textbf{\S}$\ref{S1C2}).

\subsection{Ablation Study on Different Construction Methods}
\label{S1C1}
\begin{figure}[h]
  \centering
  \begin{subfigure}[h]{0.44\linewidth}
    \centering
    \includegraphics[width=\linewidth,height=4.7cm]{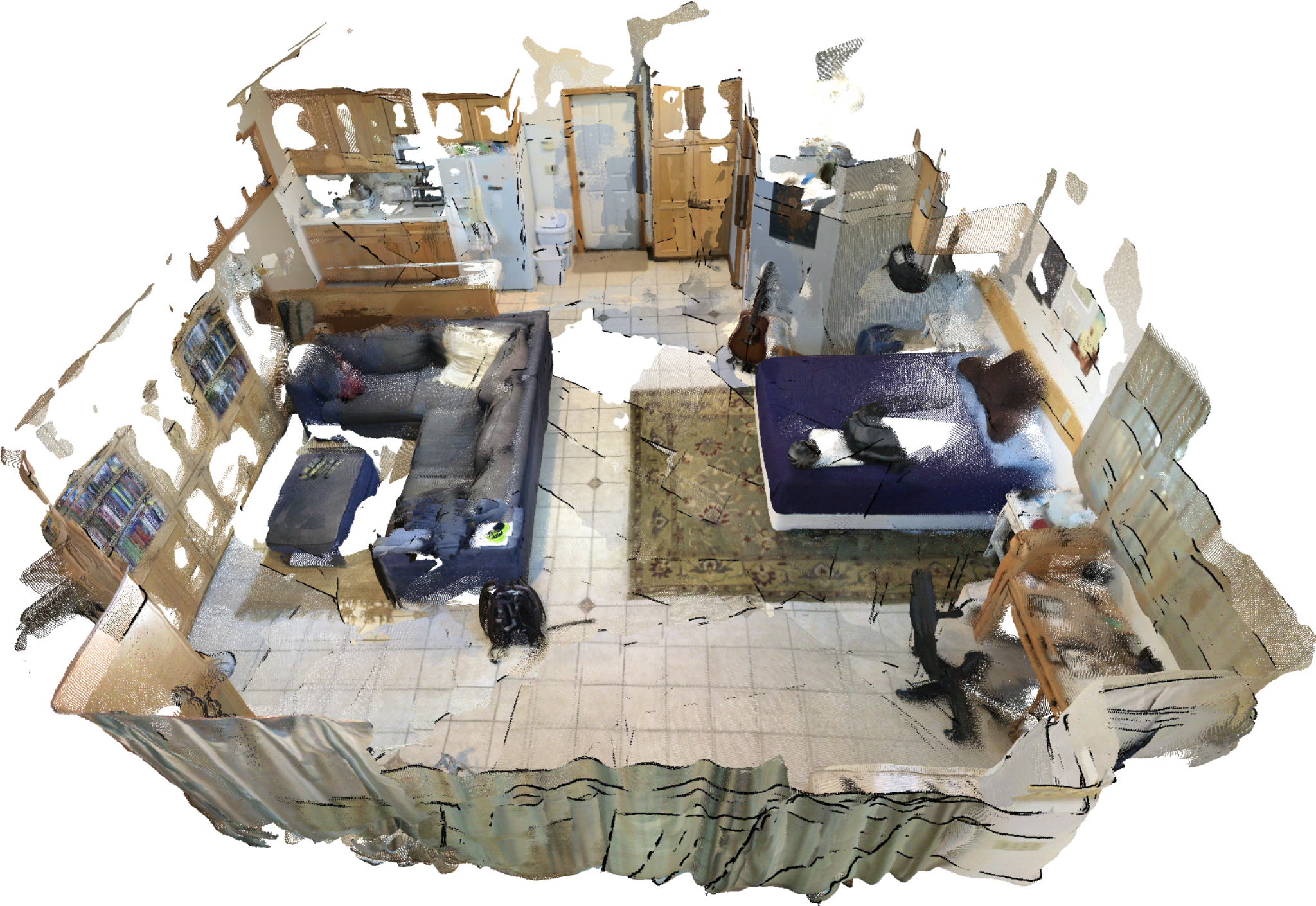}
    \caption{VGGT}
  \end{subfigure}
  \hspace{0.1\linewidth}
  \begin{subfigure}[h]{0.44\linewidth}
    \centering
    \includegraphics[width=\linewidth,height=4.7cm]{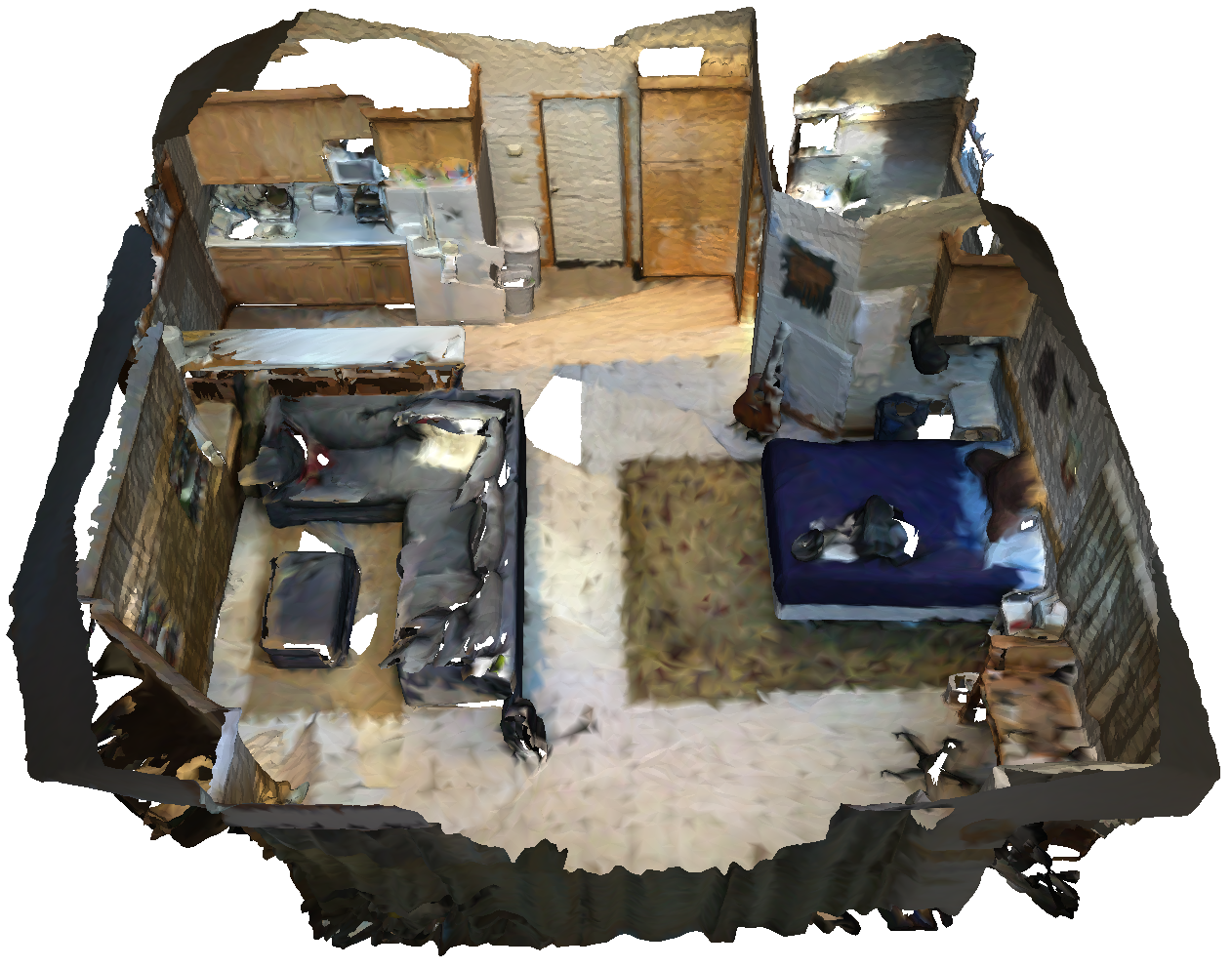}
    \caption{RGB-D (ScanNet)}
  \end{subfigure}
  \caption{Illustrative comparison of reconstructed scenes using (a) VGGT~\cite{wang2025vggt} and (b) RGB-D (ScanNet~\cite{dai2017scannet}). VGGT reconstructs scenes using only $128$ multi-view RGB images, offering convenience but lacking fine-grained details. }
  \label{fig_different_constrcution_methods}
\end{figure}

Deep learning-based 3D reconstruction methods (\textit{e.g.}, VGGT~\cite{wang2025vggt}) offer the advantage of lower computational cost, enabling direct prediction of point clouds from multi-view RGB images. In contrast, as shown in Figure~\ref{fig_different_constrcution_methods}, traditional reconstruction methods (\textit{e.g.}, ScanNet~\cite{dai2017scannet}) synthesize scene point clouds from RGB and depth (RGB-D) streams combined with camera intrinsics and extrinsics, resulting in higher accuracy and better reconstruction quality. In this work, we conduct scene parsing based on the point clouds provided by ScanNet. In future work, we will explore leveraging scenes reconstructed using VGGT to improve the efficiency of the Scene Information construction process.

\subsection{Ablation Study on Ground Truth Labels \textit{vs.} Mask3D labels}
\label{S1C2}
The type of labels usually has a notable impact on the construction of Scene Information, the generation strategy of model inputs, as well as evaluation metrics and implementation code. Following previous works, we adopt instance segmentation results generated by Mask3D~\cite{schult2023mask3d} as the default labeling method in this paper. In this section, we further investigate how using Ground Truth (GT) labels provided by the ScanNet dataset affects the final model performance, aiming to assess the influence of label quality on scene understanding effectiveness.

Table~\ref{tab_labels} presents an ablation study comparing the performance of the Text-Scene framework when using different types of instance segmentation labels. We take the text-only Scene Information as input, with Qwen2-7B~\cite{wang2024qwen2} as the base model. Across all evaluated downstream tasks, the model using GT labels weakly outperforms the one using Mask3D~\cite{schult2023mask3d} predictions. Note that for the visual grounding task, using ground truth labels eliminates bounding box prediction errors. As a result, the IoU values are either 0 or 1, leading to identical scores for both @0.25 and @0.5 thresholds in the evaluation metrics. Specifically, on the ScanRefer~\cite{chen2020scanrefer} task, GT labels lead to a noticeable improvement, raising Acc@0.25 from 59.3 to 63.5 and Acc@0.5 from 53.6 to 63.5. A similar pattern is observed in Multi3DRef~\cite{zhang2023multi3drefer}, where both F1@0.25 and F1@0.5 increase from 63.1 and 58.7 to 67.2, respectively. 

\begin{table}[h]
\centering
\caption{Ablation Study on Ground Truth labels \textit{vs.} Mask3D labels.}
\resizebox{\linewidth}{!}{
\begin{tabular}{lcc cc cc cc c}
\toprule
\multirow{2}{*}{Method} 
& \multicolumn{2}{c}{ScanRefer} 
& \multicolumn{2}{c}{Multi3DRef} 
& \multicolumn{2}{c}{Scan2Cap} 
& \multicolumn{2}{c}{ScanQA} 
& \multicolumn{1}{c}{SQA3D} \\
\cmidrule(lr){2-3} \cmidrule(lr){4-5} \cmidrule(lr){6-7} \cmidrule(lr){8-9} \cmidrule(lr){10-10}
& Acc@0.25 & Acc@0.5 & F1@0.25 & F1@0.5 & B-4@0.5 & C@0.5 & C & EM & EM \\
\midrule
Text-Scene (w/ Mask3D labels) & 59.3 & 53.6 & 63.1 & 58.7 & 36.8 & 80.0 & 89.5 & 22.9 & 57.7 \\
Text-Scene (w/ GT labels)     & \multicolumn{2}{c}{\textbf{63.5}} & \multicolumn{2}{c}{\textbf{67.2}} & \textbf{37.3} & \textbf{81.1} & \textbf{90.7} & \textbf{23.4} & \textbf{58.6} \\
\bottomrule
\end{tabular}
}
\label{tab_labels}
\end{table}

\section{Spatial Relationship Parsing Details}
\label{appendix_spatial}
Firstly, we got the bounding boxes' coordinates and size from two objects: $(x_1,y_1,z_1,w_1,h_1,l_1)$ and $(x_2,y_2,z_2,w_2,h_2,l_2)$. Then we compute the Euclidean distance $d$ between two objects as following:
\begin{equation}
    d = \sqrt{(x_1-x_2)^2+(y_1-y_2)^2+(z_1-z_2)^2}.
\end{equation}

\setlength{\intextsep}{3pt}
\begin{wrapfigure}{r}{0.46\textwidth}
    \centering
    \includegraphics[width=0.45\textwidth]{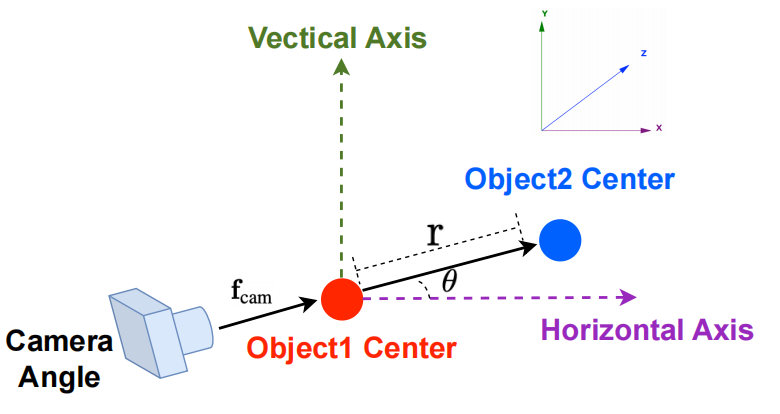}
    \caption{An example of calculating the angle between two objects. $\bm{f}_{\text{cam}}$ denotes the forward direction of the camera, while $\bm{r}$ represents the vector pointing from Object 1 toward Object 2.}
    \vspace{-5pt}
    \label{fig_camera_angle}
\end{wrapfigure}

Next, we propose a spatial relationship reasoning framework that integrates geometric proximity, camera view-based directional inference, and semantic priors to generate fine-grained and interpretable object-level spatial relations. To more robustly determine spatial proximity, we also take into account the sizes of the two objects. We calculate the maximum bounding box dimension, and define an adaptive proximity threshold based on a scaling factor $\beta$. If the distance $d$ is smaller than $\beta$ times the maximum of the two object sizes, the system classifies the spatial relation as ambiguous and assigns soft label (\textit{e.g.}, nearby). When $d$ exceeds $\beta$ times, directional reasoning is performed to determine relations such as in front of, left of, above, or an o'clock-style description. As shown in Figure~\ref{fig_camera_angle}, given the forward vector $\bm{f}_{\text{cam}}$ and the upward vector $\bm{u}_{\text{cam}}$ of the camera, we project the relative position vector $\bm{r}$ between the center of two objects onto these directional bases. We then compute the projection of $\bm{r}$ onto $\bm{u}_{\text{cam}}$ to determine whether one object is positioned above or below the other for vertical reasoning. For horizontal reasoning, we project $\bm{r}$ onto the horizontal plane orthogonal to $\bm{u}_{\text{cam}}$ and compute the deviation angle ${\theta}$ between $\bm{r}$ and $\bm{f}_{\text{cam}}$. If ${\theta}$ falls within predefined angular ranges corresponding to canonical directions (\textit{e.g.}, in front of, behind, left of, right of), we assign discrete relational labels accordingly. However, if $\theta$ deviates beyond a specified angular tolerance $\theta_{\text{tol}}$, we adopt a finer-grained o’clock-style representation. The 360-degree horizontal plane is divided into $N_{\text{o'clock}}$ equal sectors, and $\theta$ is mapped to natural language labels. In addition, we incorporate a set of semantic prior rules to account for strongly constrained object-object relationships. When a given object pair matches one of these prior templates, the semantic label from $\mathcal{R}_\text{prior}$ takes precedence, bypassing distance and angular reasoning.

\section{More Details About Self-reflaction Mechanism}
\label{S3}
\begin{figure}[b]
    \centering
    \includegraphics[width=0.95\linewidth]{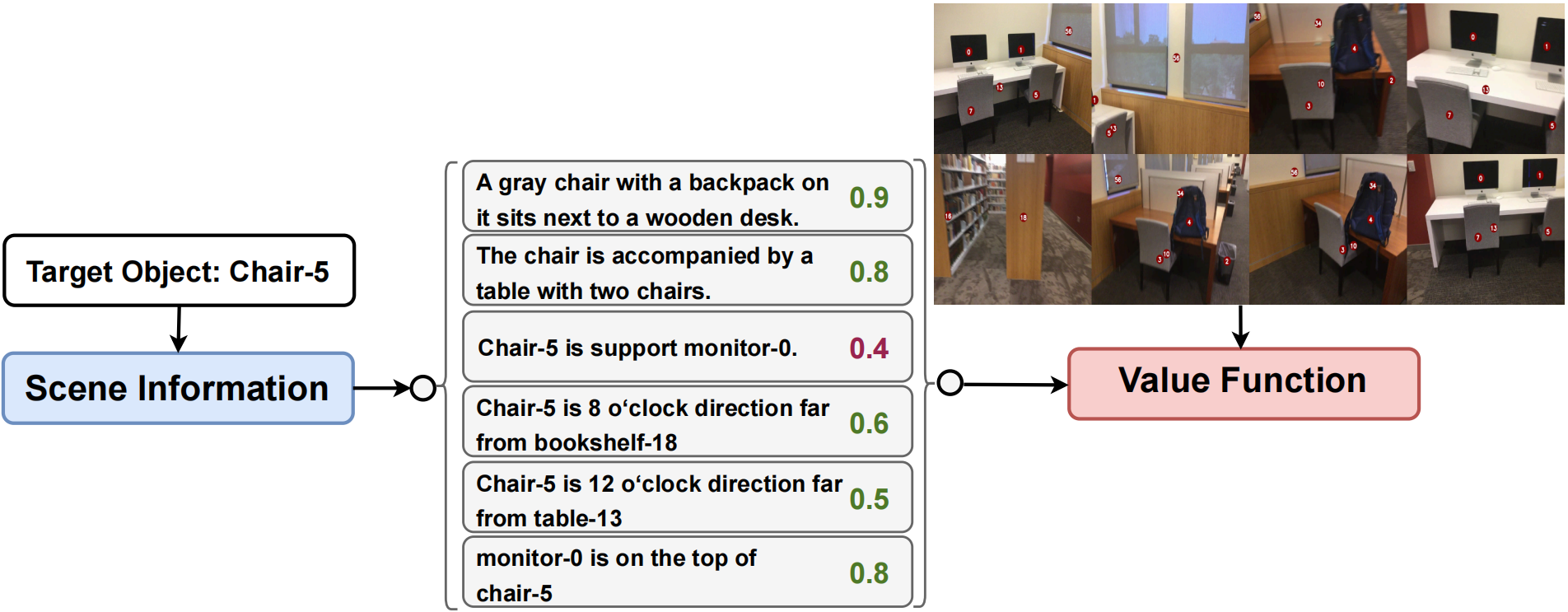}
    \caption{Illustration of the self-reflection mechanism. We mitigate hallucination issues by training a Value Function to evaluate the textual content generated by advanced LLMs, incorporating physical commonsense and human preferences into the assessment process.}
    \label{fig_self_reflaction}
\end{figure}

Figure~\ref{fig_self_reflaction} illustrates the self-reflection mechanism extensively utilized in the Text-Scene framework to enhance the quality of Scene Information. We employ a value function to evaluate the quality of Scene Information generated by a high-level LLM. This function returns a score that determines whether a given sentence should be regenerated. Our scoring model adopts the architecture of a BLIP-2~\cite{li2023blip} visual encoder and a tiny T5-3b~\cite{2020t5} model as decoder. The T5 decoder outputs a score between 0 and 1. We train the Value Function using $1,000$ samples comprising both human annotations and LLM-generated descriptions, injecting human preferences, objective facts, and physical laws. 

\section{Training Details}
\label{S5}
The training process of the LLM consists of two stages: textual scene information alignment and instruction fine-tuning on downstream tasks. All experiments were conducted using 4$\times$80G A800 GPUs with a BF16 data type. During the instruction tuning stage, we train our model for one epoch with a total batch size of 16 and a learning rate 1e-5. Throughout both stages, we employ flash-attention~\cite{dao2022flashattention}, the AdamW~\cite{loshchilov2017decoupled} optimizer, and a cosine learning rate scheduler~\cite{loshchilov2016sgdr}. Further details regarding hyper-parameters are documented in Table~\ref{tab_hyperparams_three_stage}.

\begin{table}[htbp]
\centering
\renewcommand{\arraystretch}{1} 
\setlength\tabcolsep{8pt} 
\small
\caption{Training Hyperparameters for Three Training Stages}
\resizebox{0.9\linewidth}{!}{
\begin{tabular}{ll|ll}
\toprule
\multicolumn{2}{c|}{\textbf{Text-only Fine-tuning Stage}} & \multicolumn{2}{c}{\textbf{Multimodal Fine-tuning Stage}} \\
\midrule
\textbf{Hyperparameter} & \textbf{Value} & \textbf{Hyperparameter} & \textbf{Value} \\
\midrule
 Optimizer & AdamW & Optimizer & AdamW \\
 Weight decay & 0.05 & Weight decay & 0.05 \\
 Betas & [0.9, 0.999] & Betas & [0.9, 0.999] \\
 Learning rate & $1 \times 10^{-5}$ & Learning rate & $1 \times 10^{-5}$ \\
 Warmup ratio & 0.1 & Warmup ratio & 0.1 \\
 Parallel strategy & DDP & Parallel strategy & DDP \\
 Type of GPUs & NVIDIA A800 & Type of GPUs & NVIDIA A800 \\
 Number of GPUs & 4 & Number of GPUs & 4 \\
 Batch size per GPU (total) & 4 (16) & Batch size per GPU (total) & 4 (16) \\
 Training precision & bfloat16 & Training precision & bfloat16 \\
 Gradient norm & 5.0 & Gradient norm & 5.0 \\
 Epochs & 2 & Epochs & 1 \\
 Flash Attention & \ding{51} & Flash Attention & \ding{51} \\
\bottomrule
\end{tabular}
}
\label{tab_hyperparams_three_stage}
\end{table}

\section{More Details About Proposed InPlan3D Benchmark}
\label{S4}
In this section, we provide more details about InPlan3D. Each task contains a concise high-level instruction followed by a step-by-step breakdown of low-level actions, demonstrating the following characteristic:
\begin{itemize}[leftmargin=*, topsep=0pt, itemsep=2pt, parsep=0pt]
    \item 	\textbf{Action-First Syntax}: Every step begins with a clear verb indicating the robot’s action.
	\item	\textbf{Object and Attribute References}: Actions are directed toward specific objects, referenced both semantically (e.g., “the central conference table”) and structurally (\textit{e.g.}, [table-0]).
	\item	\textbf{Spatial and Contextual Cues}: Several steps include locational qualifiers(\textit{e.g.}, beside the desk), helping localize the task in 3D space.
\end{itemize}

\begin{figure}[h]
  \centering
  \begin{subfigure}[h]{0.4\linewidth}
    \centering
    \includegraphics[width=\linewidth]{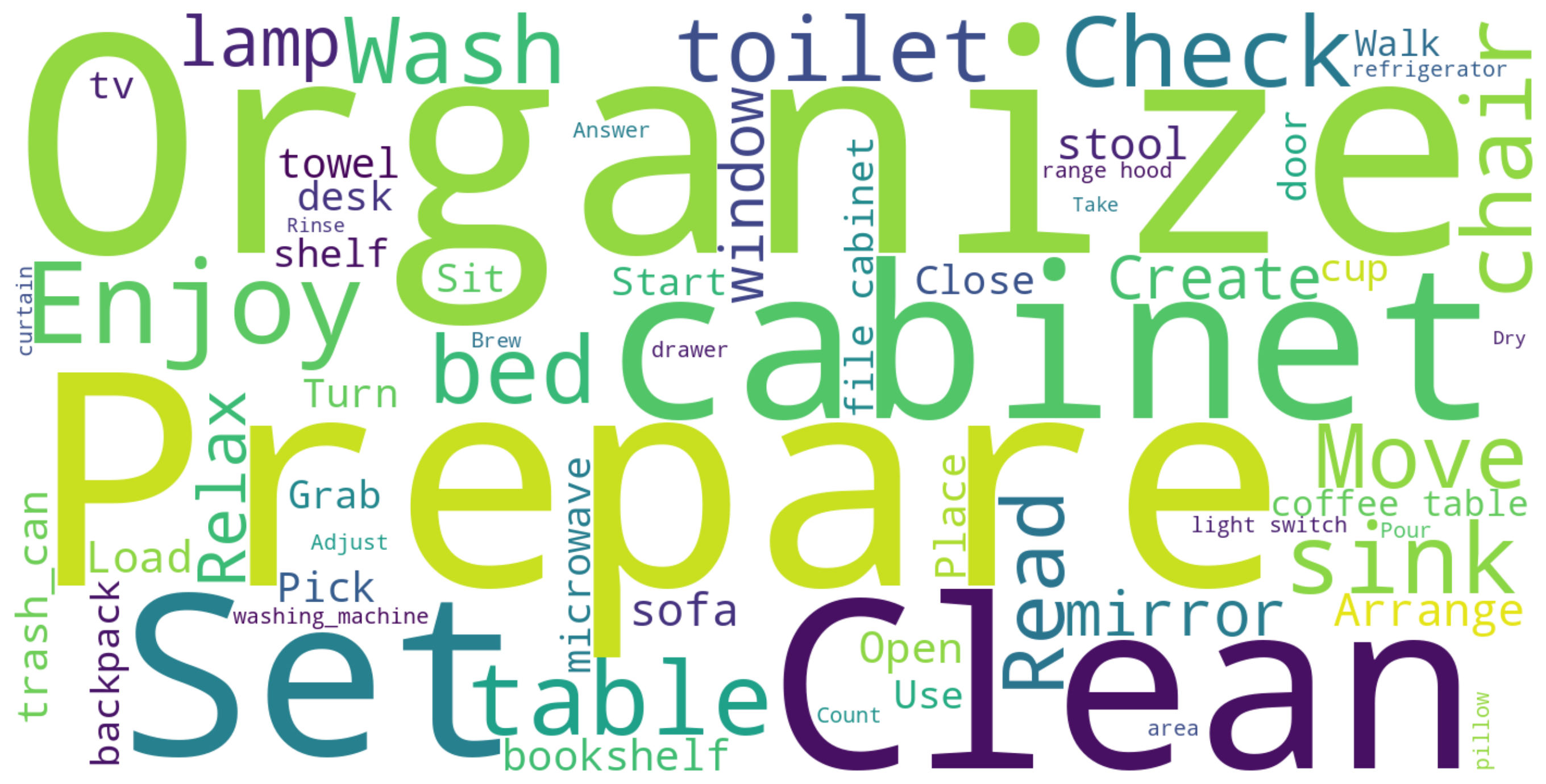}
    \caption{Actions}
  \end{subfigure}
  \hspace{0.02\linewidth}
  \begin{subfigure}[h]{0.4\linewidth}
    \centering
    \includegraphics[width=\linewidth]{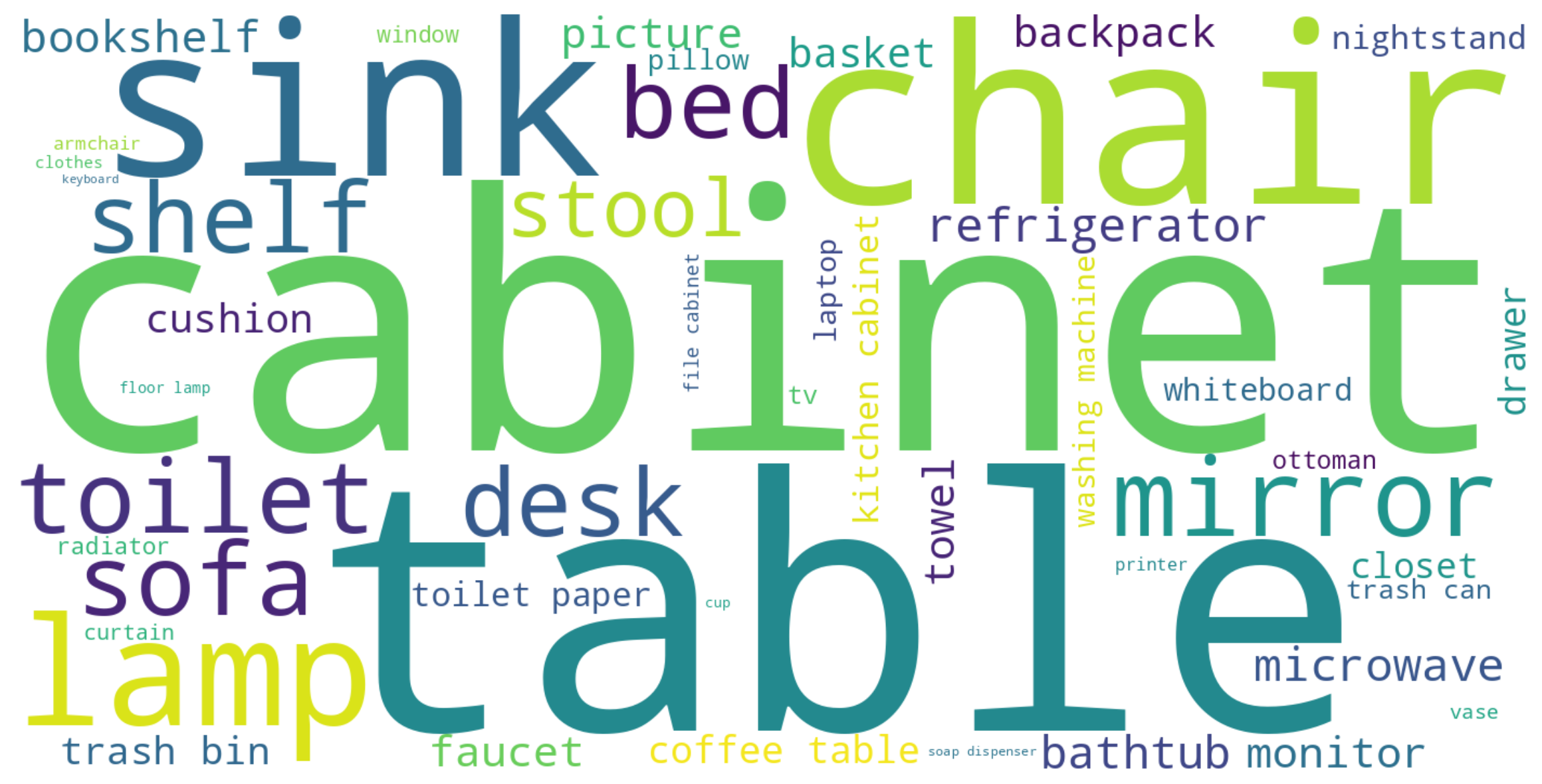}
    \caption{Objects}
  \end{subfigure}
  \caption{Wordclouds of (a) Actions and (b) Objects}
  \label{fig_wordclouds}
\end{figure}

\begin{figure}[h]
  \centering
  \begin{subfigure}[h]{0.45\linewidth}
    \centering
    \includegraphics[width=\linewidth]{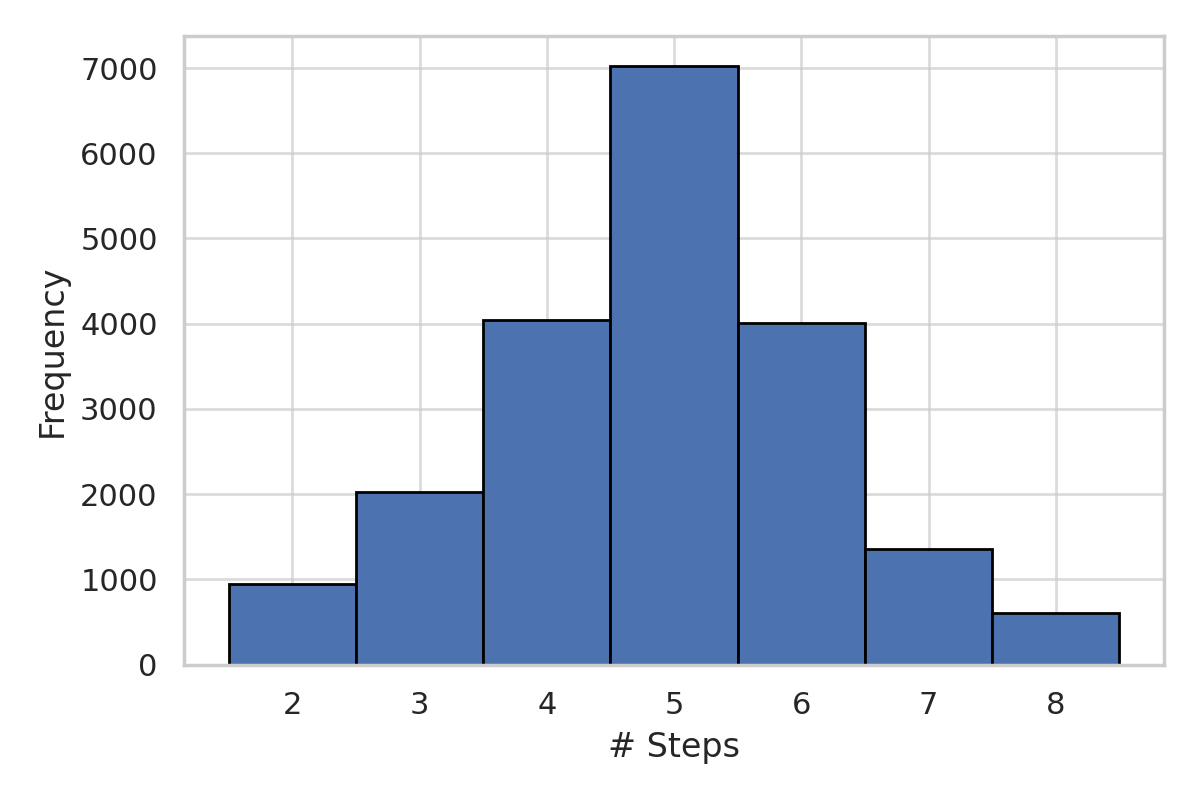}
    \caption{Steps Number Distribution}
  \end{subfigure}
  \hspace{0.02\linewidth}
  \begin{subfigure}[h]{0.45\linewidth}
    \centering
    \includegraphics[width=\linewidth]{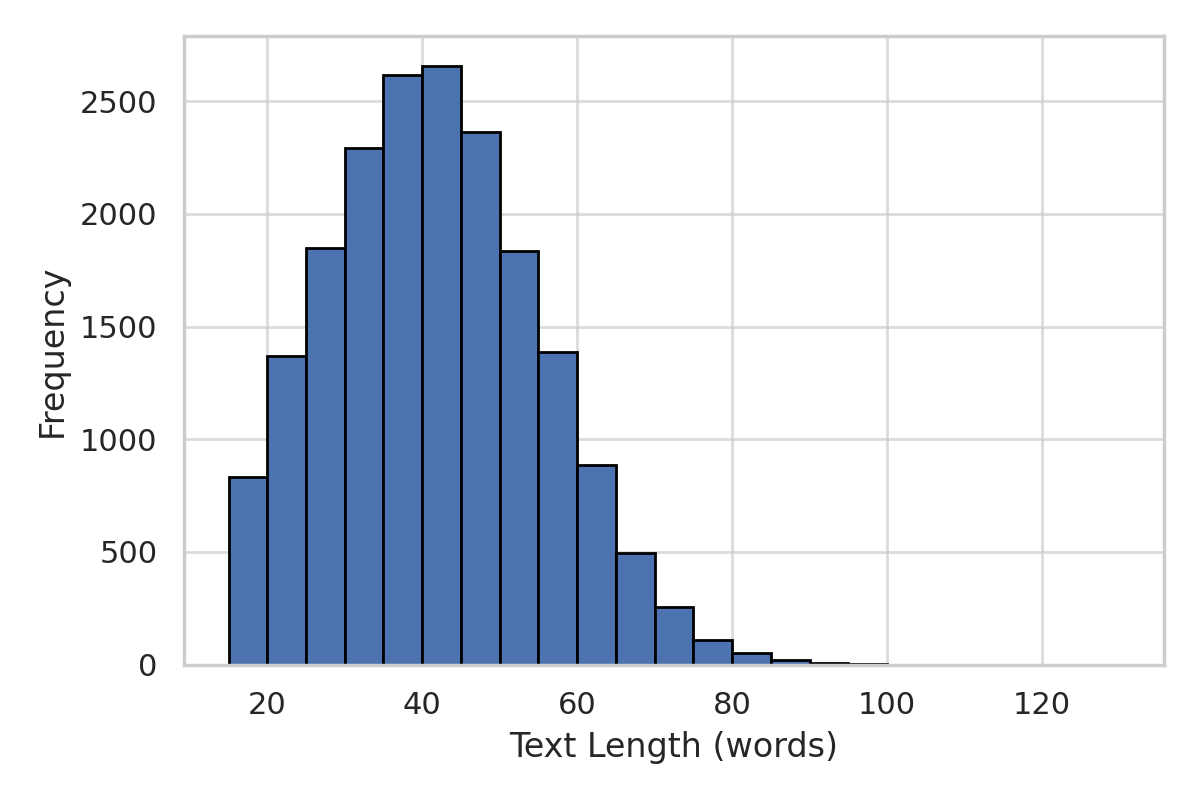}
    \caption{Text Length Distribution}
    \label{fig:wordcloud-b}
  \end{subfigure}
  \caption{Data Distribution of (a) Steps Number and (b) Text Length}
  \label{fig_data_distribution}
\end{figure}

\begin{figure}[h]
    \centering
    \includegraphics[width=\linewidth]{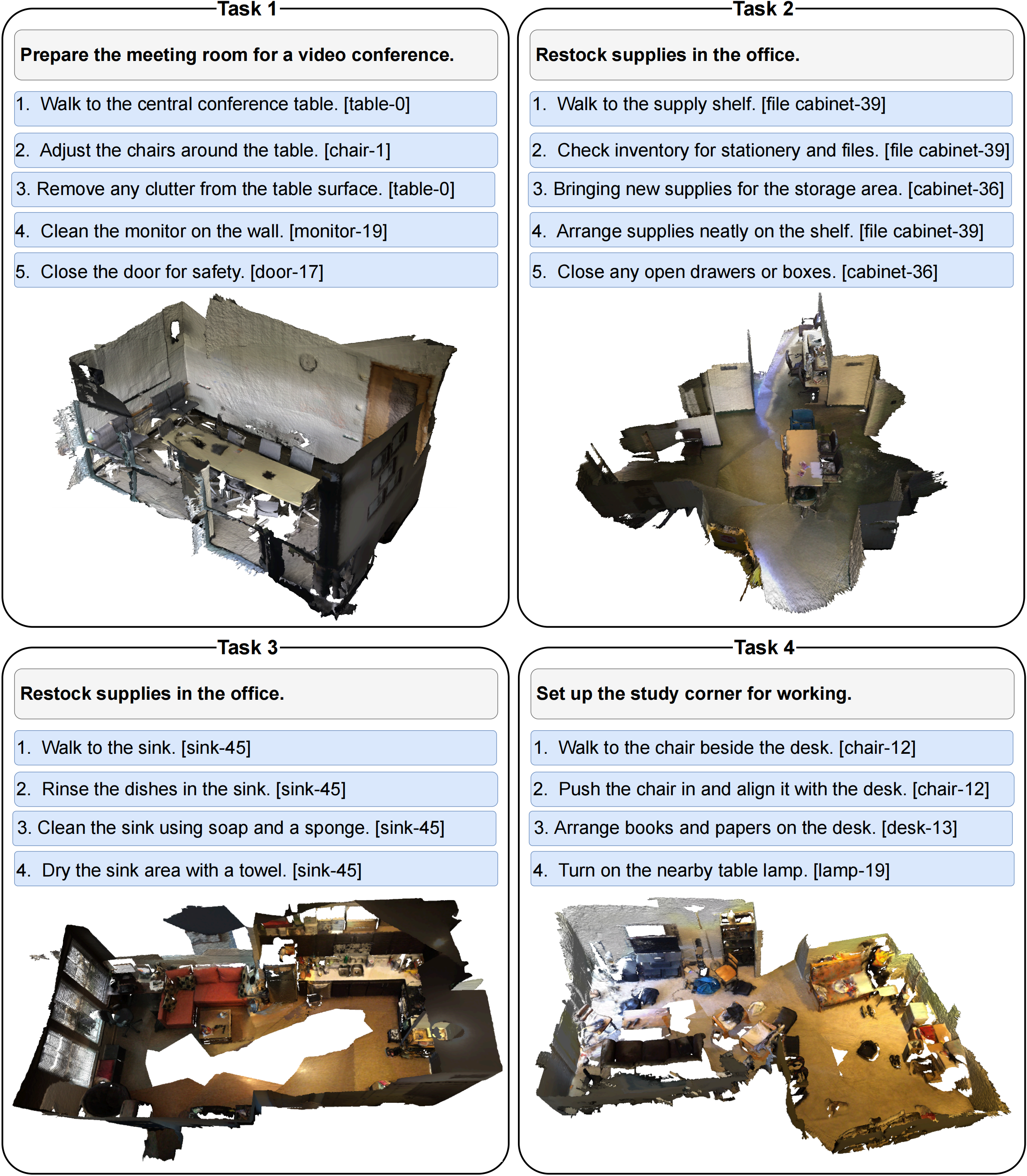}
    \caption{Examples of InPlan3D benchmark. }
    \label{fig_examples_planning}
\end{figure}

In Figure~\ref{fig_wordclouds}, we provide word cloud analyses of the Actions and Objects appearing in the dataset. As shown in Figure~\ref{fig_data_distribution}, most tasks in InPlan3D contain $4$ to $6$ steps with $30$ to $60$ words in total. Figure~\ref{fig_examples_planning} presents several examples from InPlan3D.

\vspace{-10pt}

\section{Prompts for Advanced Closed-source MLLMs}
\begin{figure}[htbp]
    \centering
    \includegraphics[width=\linewidth]{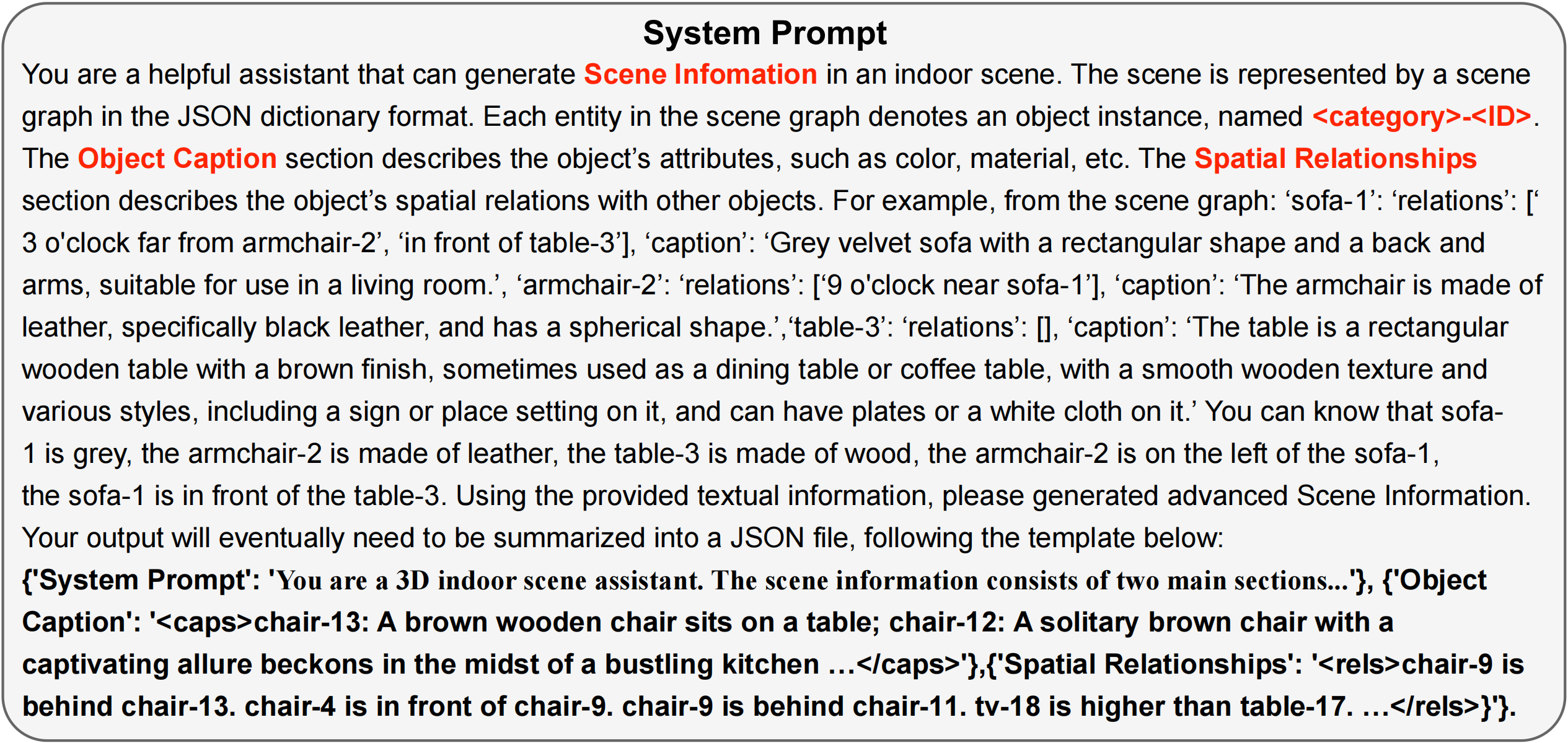}
    \caption{Prompts for constructing Scene Information. We show the prompts used for GPT-4o, which consists of \textit{System Prompt}, \textit{Object Caption} and \textit{Spatial Relationships}. After generating the response, we further refine the content by self-reflection mechanism.}
    \label{fig_prompt_scene information}
\end{figure}

\begin{figure}[htbp]
    \centering
    \includegraphics[width=\linewidth]{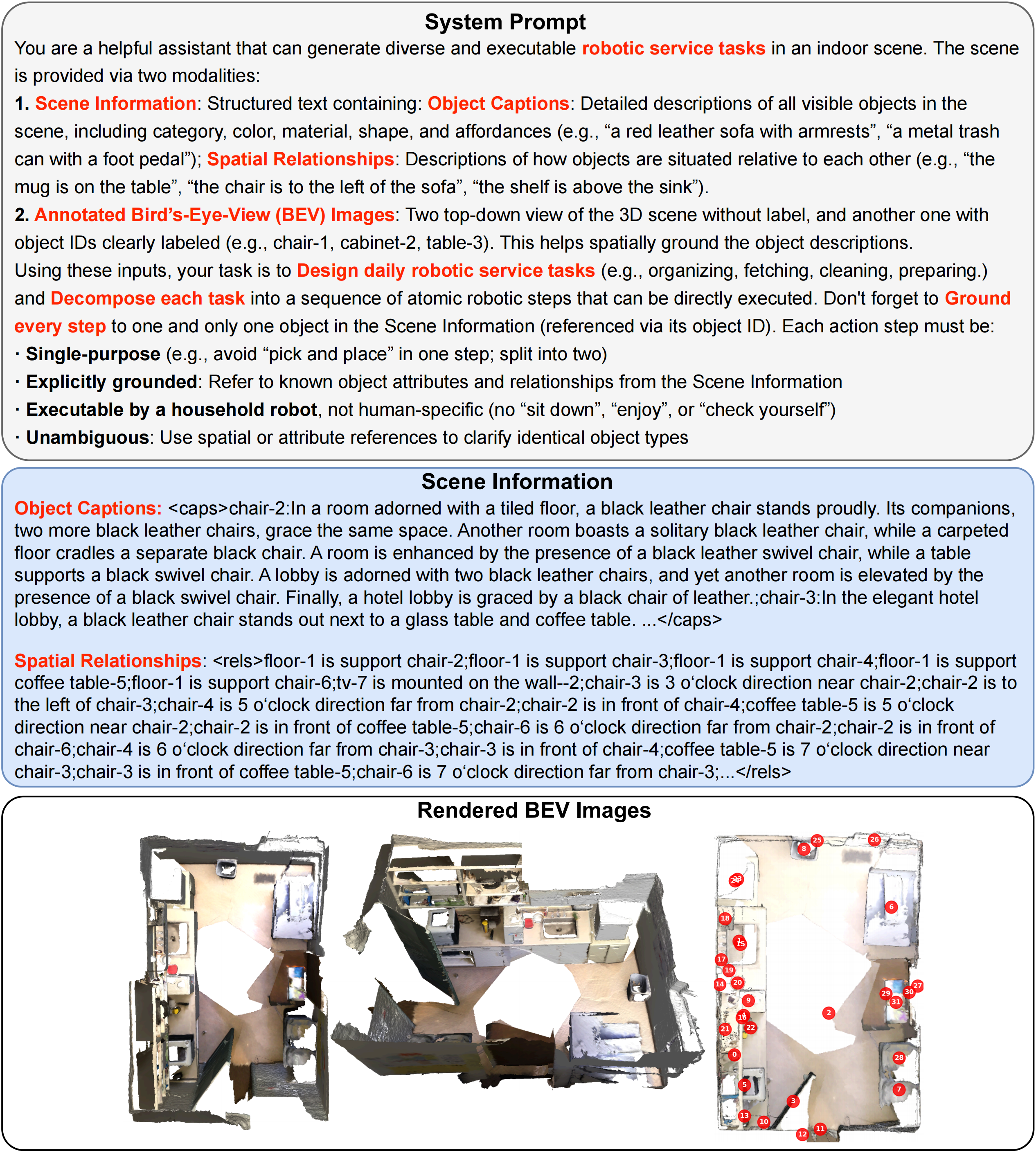}
    \caption{Prompts for constructing InPlan3D benchmark. We show the prompts used for GPT-4o, which consists of \textit{System Prompt}, \textit{Scene Information} and \textit{Rendered BEV Images}. After generating the response, we further refine the content to ensure it conforms to the predefined format.}
    \label{fig_prompt_planning}
\end{figure}

In this section, we provide examples of the prompts used for generating the Scene Information (refer to Figure~\ref{fig_prompt_scene information}) and the InPlan3D dataset (refer to Figure~\ref{fig_prompt_planning}). 

\end{document}